\documentclass{article}

\usepackage{microtype}
\usepackage{graphicx}
\usepackage{subfigure}
\usepackage{booktabs} % for professional tables

\usepackage{hyperref}

\usepackage[dvipsnames]{xcolor}
\usepackage{arydshln}

\renewenvironment{thebibliography}[1]{
  \begin{oldthebibliography}{#1}
    \setlength{\itemsep}{-5em}
    \setlength{\parskip}{-5em}
}
{
  \end{oldthebibliography}
}

\newcommand{\cyan}[1]{{\color{cyan}#1}}
\newcommand{\blue}[1]{{\color{blue}#1}}
\newcommand{\red}[1]{{\color{red}#1}}

\newcommand{\BPA}[1]{{\fontsize{#1}{#1}\selectfont\textbf{BPA}}}
\newcommand{\BPAsml}{\BPA{8.1}}
\newcommand{\BPAreg}{\BPA{10}}
\newcommand{\BPAps}{{\BPAsml\color{blue}{$_\textit{\textbf{p}}$}}}
\newcommand{\BPAis}{{\BPAsml\color{cyan}{$_\textit{\textbf{i}}$}}}
\newcommand{\BPAts}{{\BPAsml\color{red}{$_\textit{\textbf{t}}$}}}

\newcommand{\BPAp}{{\BPAreg\color{blue}{$_\textit{\textbf{p}}$}}}
\newcommand{\BPAi}{{\BPAreg\color{cyan}{$_\textit{\textbf{i}}$}}}
\newcommand{\BPAt}{{\BPAreg\color{red}{$_\textit{\textbf{t}}$}}}

\usepackage[accepted]{icml2024}

\usepackage{amsmath}
\usepackage{amssymb}
\usepackage{mathtools}
\usepackage{amsthm}

\DeclareMathOperator*{\argminA}{arg\,min} % Jan Hlavacek

\usepackage[capitalize,noabbrev]{cleveref}

\theoremstyle{plain}

\theoremstyle{definition}

\theoremstyle{remark}

\usepackage[textsize=tiny]{todonotes}

\icmltitlerunning{The Balanced-Pairwise-Affinities Feature Transform}

\begin{document}

\twocolumn[
\icmltitle{The Balanced-Pairwise-Affinities Feature Transform}

% It is OKAY to include author information, even for blind
% submissions: the style file will automatically remove it for you
% unless you've provided the [accepted] option to the icml2024
% package.

% List of affiliations: The first argument should be a (short)
% identifier you will use later to specify author affiliations
% Academic affiliations should list Department, University, City, Region, Country
% Industry affiliations should list Company, City, Region, Country

% You can specify symbols, otherwise they are numbered in order.
% Ideally, you should not use this facility. Affiliations will be numbered
% in order of appearance and this is the preferred way.
%\icmlsetsymbol{equal}{*}

\begin{icmlauthorlist}
\icmlauthor{Daniel Shalam}{uoh}
\icmlauthor{Simon Korman}{uoh}
\end{icmlauthorlist}

\icmlaffiliation{uoh}{Department of Computer Science, University of Haifa, Israel}

\icmlcorrespondingauthor{Daniel Shalam}{dani360@gmail.com}
%\icmlcorrespondingauthor{Simon Korman}{skorman@cs.haifa.ac.il}

% You may provide any keywords that you
% find helpful for describing your paper; these are used to populate
% the "keywords" metadata in the PDF but will not be shown in the document
\icmlkeywords{Machine Learning, ICML}

\vskip 0.3in
]

% this must go after the closing bracket ] following \twocolumn[ ...

% This command actually creates the footnote in the first column
% listing the affiliations and the copyright notice.
% The command takes one argument, which is text to display at the start of the footnote.
% The \icmlEqualContribution command is standard text for equal contribution.
% Remove it (just {}) if you do not need this facility.

\printAffiliationsAndNotice{}  % leave blank if no need to mention equal contribution
% \printAffiliationsAndNotice{\icmlEqualContribution} % otherwise use the standard text.

%-------------------------------------------------------------------
% abstract
%-------------------------------------------------------------------
\begin{abstract}

The Balanced-Pairwise-Affinities (BPA) feature transform is designed to upgrade the features of a set of input items to facilitate downstream matching or grouping related tasks. 
The transformed set encodes a rich representation of high order relations between the input features. 
A particular min-cost-max-flow fractional matching problem, whose entropy regularized version can be approximated by an optimal transport (OT) optimization, leads to a transform which is efficient, differentiable, equivariant, parameterless and probabilistically interpretable.
While the Sinkhorn OT solver has been adapted extensively in many contexts, we use it differently by minimizing the cost between a set of features to \textit{itself} and using the transport plan's \textit{rows} as the new representation.
Empirically, the transform is highly effective and flexible in its use and consistently improves networks it is inserted into, in a variety of tasks and training schemes. 
We demonstrate state-of-the-art results in few-shot classification, unsupervised image clustering and person re-identification. 
Code is available at \url{github.com/DanielShalam/BPA} .
\end{abstract}

% {
% We introduce the Balanced-Pairwise-Affinities (BPA) feature transform which is designed to upgrade the set of features of a data instance to be highly effective with respect to downstream tasks that rely on matching or grouping operations. 

% The transformed set encodes a rich representation of high order relations between the instance features. A simple distance computation between a pair of transformed features captures the \textit{direct} similarity between the original features, as well as their \textit{third party} `agreement' regarding the other features in the set. 

% We pose a particular fractional matching problem, modeled as a min-cost-max-flow instance, whose entropy regularized version can be approximated by an optimal transport (OT) based optimization.

% The OT based solution gives rise to our transform which is fast to compute, differentiable, equi-variant, parameter-less and whose output has a clear probabilistic interpretation with respect to feature similarities.

% We show empirically, that the suggested transform is highly effective and flexible in its use, consistently improving networks that it is inserted into, in a variety of tasks and training schemes. We demonstrate its merits through the problem of unsupervised clustering, and show its efficiency and wide applicability in both few-shot-classification, for which we obtain state-of-the-art results, as well as in large-scale person re-identification tasks. 
% } % <<<<<<<<<<<<<<<<<<<<<<<<<

%-------------------------------------------------------------------
% introduction
%-------------------------------------------------------------------
\vspace{-9pt}
\section{Introduction}

%--------------------------------------------------------------------

In this work, we reassess the functionality of features in \textit{set-input} problems, in which a task is defined over a \textit{set} of items. Prominent examples of this setting are few-shot classification {\cite{ravi2017optimization}}, clustering {\cite{van2020scan}}, feature matching {\cite{korman2015coherency}} and person re-identification {\cite{ye2021deep}}, to name but a few. In such tasks, features computed at test time are mainly compared relative to one another, and less so to the features seen at training time. For such tasks, the practice of learning a generic feature extractor during training and applying it at test time is sub-optimal. 

%The main challenge is in designing training and inference schemes, is in finding ways to exploit large corpuses of training data to learn features that can easily adapt, or be relevant, to the test time task. 
%
%Our approach to doing so will be in the form of a \textit{feature transform} that jointly re-embeds the set of features of an instance in a way that resembles how recently popular self-attention mechanisms and Transformers \cite{ramachandran2019stand,lee2019set,mialon2021trainable,khan2021transformers} re-embed sets of features.
%
%Being at the low-to-mid level of the relevant pipelines for set-input problems, advances in representation learning in this setting could have direct impact and wide applicability in many related applications. 
%
%The general idea of the Balanced-Pairwise-Affinities (BPA) feature transform that we propose is depicted and explained in Fig. \ref{fig.BPA_illustration}, as part of the general design of networks that work on sets,
%(\simon{sets? IS?}), 
%illustrated in Fig. \ref{fig.network}.

%% -------------------- figure Generic NETWORKs ---------------
\begin{figure*}[t]
\centering \vspace{-6pt} %\hspace{-27pt}
% \begin{tabular}{c c}
%      \includegraphics[height=0.305\textwidth]{figures/network.png} 
%      &  \includegraphics[height=0.305\textwidth]{figures/network_T.png} \\
% \end{tabular}
\hspace{9pt}
\includegraphics[width=0.99\textwidth]{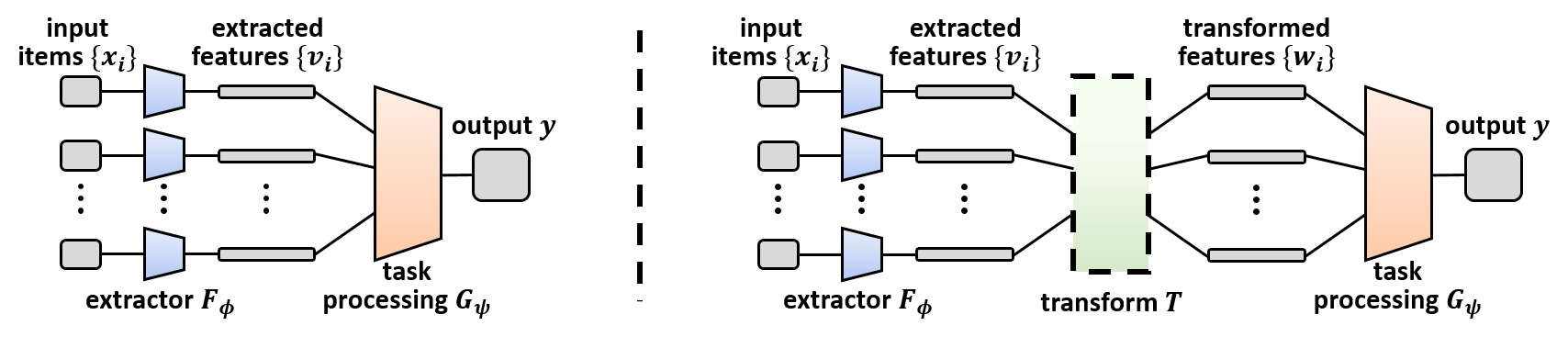}
   \vspace{-4pt}\caption{
   \textbf{Generic designs of networks that act on \textit{sets} of items.} These cover relevant architectures, e.g. for few-shot-classification and clustering. \textbf{Left:} A generic network for processing a set of input items typically follows the depicted structure: (i) Each item \textit{separately} goes through a common feature extractor $F$. (ii) The set of extracted features is the input to a downstream task processing module $G$. ; \textbf{Right:} A more general structure in which the extracted features undergo a \textit{joint} processing by a transform $T$. Our BPA transform (as well as other attention mechanisms) is of this type and its high-level design (within the `green' module) is detailed in Fig. \ref{fig.BPA_illustration}.}
    \label{fig.network}\vspace{-10pt}
\end{figure*}

In set-input problems, such as few-shot classification, an \textit{instance} of the task is in the form of a set of $n$ items (e.g. images) $\{x_i\}_{i=1}^n$.
A generic neural-network pipeline (Fig. \ref{fig.network} Left) typically uses a feature embedding (extractor) $F$, that is applied independently to each input item, to obtain a set of features ${V}\hspace{-2pt}=\hspace{-2pt}\{v_i\}_{i=1}^n\hspace{-2pt}=\hspace{-2pt}\{F(x_i)\}_{i=1}^n$, prior to downstream task-specific processing $G$ (e.g. a clustering head or classifier).
The features ${V}$ can be of high quality (concise, unique, descriptive), but are limited in representation since they are extracted based on knowledge acquired for similar examples at train time, with no context of the test time instance they are part of, which is critical in set-input tasks. 

We rather consider the more general framework (Fig. \ref{fig.network} Right), in which the per-item independently extracted feature collection $V$ is passed to an attention-mechanism type computation, in which some \textit{transform} jointly processes the entire set of instance features, re-embedding each feature in light of the joint statistics of the entire instance. 

The main idea of BPA is very intuitive and is demonstrated on a toy example in Fig. \ref{fig.BPA_illustration}. The embedding of each feature will encode the \textit{distribution} of its affinities to the rest of the set items. Specifically, items in the embedded space will be close if and only if they share a similar such distribution, i.e. 'agree' on the way they 'see' the entire set. 
In fact, the transform largely discards the item-specific feature information, resulting in a purely relative normalized representation that results in a highly efficient embedding with many attractive properties. %In other words, the embedding of each item depends mainly on the other items of the set (and will be transformed differently when belonging to different instances).

The proposed transformation can be computed very efficiently, with negligible runtime within the hosting network, and can be easily used in different contexts, as can be seen in the pseudo-code snippets we provide in Sections \ref{sec:BPA_impl} and \ref{sec:insertions} of the Appendix. 
The embedding itself is given by rows of an optimal-transport (OT) plan matrix, which is the solution to a regularized min-cost-max-flow fractional matching problem that is defined over the pairwise (self)-affinities matrix of the features in the set. 

Technically, it involves the computation of pairwise distances  and several normalization iterations of a Sinkhorn \cite{cuturi2013sinkhorn} algorithm, baring apparent similarities to many related methods based on either Spectral Clustering \cite{ng2001spectral} that normalize the same affinity matrix), attention-mechanisms \cite{vaswani2017attention} that learn features based on a self-affinities matrix perhaps even normalized \cite{sander2022sinkformers} and other matching \cite{sarlin2020superglue} or classification \cite{hu2020leveraging} algorithms where optimal-transport plans are computed between source items and target items or class centers.
However, the most important difference and our main novel observation is that \underline{the self fractional matching itself} (which can be viewed as a balanced affinity matrix) can serve as a powerful embedding, since the distances in this space (between assignment vectors) have explicit interpretations that we explore, which are highly beneficial to general grouping based algorithms that are applied to such set-input tasks.

%The proposed transform can be seen as a special case of an attention mechanism \cite{ramachandran2019stand} specialized to features of set-input tasks, with required adaptations. Techniques developed here borrow from and might lend to those used in  
    % set-to-set \cite{zaheer2017deep,ye2020few,maron2020learning}, 
    % self-attention \cite{ramachandran2019stand,mialon2021trainable}
    % and transformer \cite{lee2019set,khan2021transformers}
    % architectures.

%--------------------------------------------------
\vspace{-1pt}
\subsection*{Contribution}\vspace{-4pt}
%--------------------------------------------------

We propose a parameter-less optimal-transport based feature transform, termed BPA, which can be used as a drop-in addition that converts a generic feature extraction scheme to one that is well suited to set-input tasks (e.g. from \cref{fig.network} Left to Right). 
It is analyzed and shown to have the following attractive set of qualities. 
(i) \textit{efficiency}: having real-time inference; 
(ii) \textit{differentiability}: allowing end-to-end training of the entire `embedding-transform-inference' pipeline of  Fig. \ref{fig.network} Right; 
(iii) \textit{equivariance}: ensuring that the embedding works coherently under any order of the input items; 
(iv) \textit{probabilistic interpretation}: 
each embedded feature will encode its distribution of affinities to all other features, by conforming to a doubly-stochastic constraint;
(iv) \textit{valuable metrics for the item set}: 
Distances between embedded vectors will include both direct and indirect (third-party) similarity information between input features.
%; 
%(vi) \textit{instance-aware dimensionality}: embedding dimension (capacity) is adaptive to input size (complexity).

Empirically, we show BPA's flexibility and ease of application to a wide variety of tasks, by incorporating it in leading methods of each type. We test different configurations, such as whether the hosting network is pre-trained or re-trained with BPA inside, across different backbones, whether transductive or inductive.
%A controlled experiment on unsupervised clustering is used to verify its performance, with a detailed analysis. 
Few-shot-classification is our main application with extensive experimentation on standard benchmarks, testing on unsupervised-image-clustering shows the potential of BPA in the unsupervised domain and the person-re-identification experiments show how BPA deals with non-curated large-scale tasks.
In all three applications, over the different setups and datasets, BPA consistently improves its hosting methods, achieving new state-of-the-art results.

\section{Relation to Prior Work}
%--------------------------------------------------------------------

% \newcommand{\etal}{\textit{et al.}}
\subsection{Related Techniques}
% ~~~~~~~~~~~~~~~~~~~~~~~~~~~~~~~~~~
% ~~~~~~~~~~~~~~~~~~~~~~~~~~~~~~~
\paragraph{Set-to-Set (or Set-to-Feature) Functions}
\hspace{-9pt}
have been developed to act jointly on a set of items (typically features) and output an updated set (or a single feature), which are used for downstream inference tasks.
Deep-Sets~\cite{zaheer2017deep}  formalized fundamental requirements from architectures that process sets. Point-Net~\cite{qi2017pointnet} presented an influential design for learning local and global features on 3D point-clouds, while \citet{maron2020learning} study the design of equi/in-variant layers. 
Unlike BPA, the joint processing in these methods is limited, amounting to weight-sharing between separate processes and joint aggregations. 
\vspace{-7pt}
\paragraph{Attention Mechanisms.}
% ~~~~~~~~~~~~~~~~~~~~~~~~~~~~~~~
\hspace{-9pt}
The introduction of Relational Networks \cite{santoro2017simple} and Transformers~\cite{vaswani2017attention} with their initial applications in vision models~\cite{ramachandran2019stand} have lead to the huge impact of Vision Transformers (ViTs)~\cite{dosovitskiy2020image} in many vision tasks \cite{khan2021transformers}. 
While BPA can be seen as a self-attention module, it is very different, first, since it is \textit{parameterless}, and hence can work at test-time on a pre-trained network. In addition, is can provide an explicit probabilistic global interpretation of the instance data. 
%
%Several attention based methods are designed \hl{for few-shot-learning}, such as ReNet~\cite{ReNet}, DeepEMD~\cite{DeepEMD} and FEAT~\cite{ye2020few}.
%, which we survey in the respective section.

%% -------------------- figure BPA TRANSFORM ---------------
\begin{figure*}[t]
\centering\vspace{-2pt}
    \includegraphics[width=0.98\textwidth]{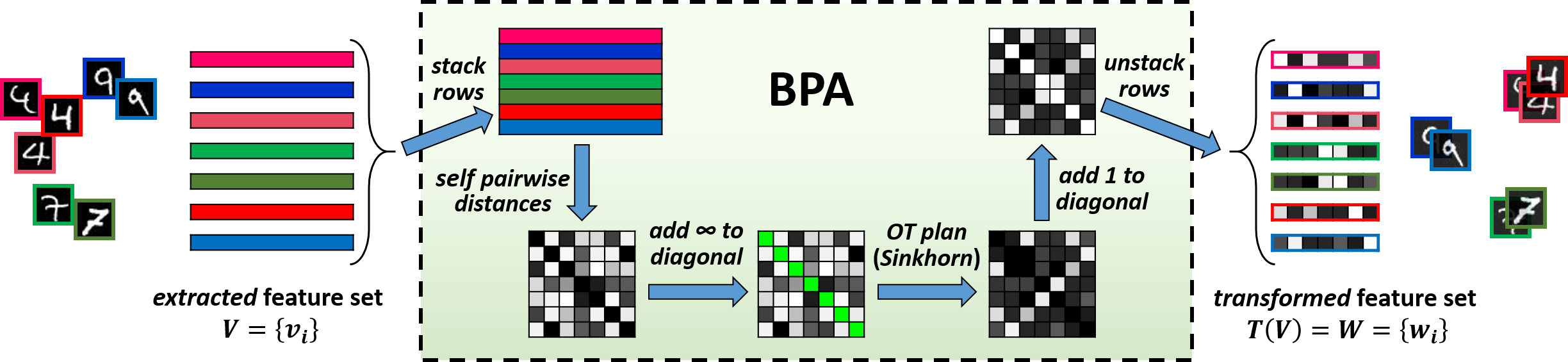}  
   \vspace{-5pt}
   \caption{    {%\fontsize{8.5}{8.5} \selectfont
   \textbf{The BPA transform:} illustrated on a toy 7 image 3-class
MNIST example.}   }    \label{fig.BPA_illustration}\vspace{-10pt}
\end{figure*}
%% ---------------------------------------------------

\vspace{-7pt}
\paragraph{Spectral Methods}
% ~~~~~~~~~~~~~~~~~~~~~~~~~~
\hspace{-9pt}
have been widely used as simple transforms applied on data that needs to undergo grouping or search based operations, jointly processing the set of features, resulting in a compact and perhaps discriminative representation. 
PCA \cite{pearson1901liii} provides a joint dimension reduction, which maximally preserves data variance, but does not necessarily improve feature affinities for downstream tasks.
Spectral Clustering (SC) ~\cite{shi2000normalized, ng2001spectral} is the leading non-learnable clustering method in use in the field. 
If we ignore its final clustering stage, SC consists of 
% (i) 
forming a pairwise affinity matrix which is 
% (ii)
normalized \cite{zass2006doubly} before 
% (iii) 
extracting its leading eigenvectors, which form the final embedding.
BPA is also based on normalizing an affinity matrix, but uses this matrix's rows as embedded features and avoids any further spectral decompositions, which are costly and difficult to differentiate through. 
%SpectralNet~\cite{shaham2018spectral} approximates spectral clustering, using a neural-net, but it needs to be trained on each particular clustering instance to be solved.

\vspace{-7pt}
\paragraph{{Optimal Transport} (OT)}
% ~~~~~~~~~~~~~~~~~~~~~~~~~~~~~~~
\hspace{-8pt}
%\noindent \textbf{++optimal-transport}\\
problems are directly related to measuring distances between distributions or sets of features. 
\citet{cuturi2013sinkhorn} popularized the Sinkhorn algorithm which is a simple, differentiable and fast approximation of entropy-regularized OT, which has since been used extensively, for 
clustering \cite{lee2019set, asano20self}, % Set transformer~
few-show-classification \cite{huang2019few,ziko2020laplacian,hu2020leveraging,zhang2021sill,chen2021few,zhu2022ease},
matching \cite{wang2019learning, fey2020deep, sarlin2020superglue}, % 2 x deep graph matching
representation learning \cite {caron2020unsupervised, asano20self},
retrieval \cite{xie2020differentiable},
person re-identification \cite{wang2022optimal},
style-transfer \cite{kolkin2019style}
and
attention~\cite{sander2022sinkformers}.
% The Set transformer~\cite{lee2019set} uses an OT-based clustering algorithm, SuperGlue~\cite{sarlin2020superglue} uses OT in an end-to-end manner for feature-point matching, and many few-shot learning methods, which we review next, have adopted the Sinkhorn algorithm to model relations between features and class representations. 
% The differentiability and efficiency of regularized OT has been shown useful to derive a differentiable `top-k' operator \cite{xie2020differentiable} or for style transfer applications \cite{kolkin2019style}. 
% Recently, Sander \etal introduced Sinkformers~\cite{sander2022sinkformers}, which are transformers in which the soft-max operation between queries and keys is replaced by Sinkhorn iterations in order to enforce double-stochasticity on the attention matrices. 
%%

Our approach also builds on some attractive properties of the Sinkhorn solver.
While our usage of Sinkhorn is extremely simple (see \cref{algo:BPA}), it is fundamentally different from all other OT usages we are aware of, since: 
(i) We compute the transport-plan between a set of features and \textit{itself} - not between feature-sets and label/class-prototypes 
\cite{huang2019few,ziko2020laplacian,hu2020leveraging,zhang2021sill,chen2021few,zhu2022ease, lee2019set, asano20self, xie2020differentiable, wang2022optimal, kolkin2019style},
or between two different feature-sets 
\cite{wang2019learning, fey2020deep, sarlin2020superglue, sander2022sinkformers};
(ii) While others use the transport-plan to obtain distances or associations between features and features/classes, we use \textit{its own rows} as new feature vectors for downstream tasks.

%we focus a \textit{self} application of OT, which enables concise modelings of the relative similarities within a set of items, and use its output transport plan \textit{directly} as a set of features.

% ==========================================
\subsection{Instance-Specific Applications}
% ==========================================
  
\vspace{-0pt}
\paragraph{Few-Shot Classification (FSC)}
% ~~~~~~~~~~~~~~~~~~~~~~~~~~~~~~~~~~
% few shot
\hspace{-8pt}
%\cite{vinyals2016matching} 
is a branch of few-shot learning in which a classifier learns to recognize previously unseen classes given a limited number of labeled examples. 
%A FSC task~\cite{vinyals2016matching}  is a self-contained instance that includes both support (labeled) and query (unlabeled) items - a clear instance-specific setup which BPA can handle.

%% meta-learning
In the \textit{meta-learning} approach, the training data is split into tasks (or episodes) mimicking the test time tasks to which the learner is required to generalize. 
% MAML and Proto
MAML~\cite{finn2017model} ``learns to fine-tune" by learning a network initialization from which it can quickly adapt to novel classes.
In ProtoNet~\cite{snell2017prototypical}, a learner is meta-trained to predict query feature classes, based on distances from support class-prototypes in the embedding space. 
% differentiation
The trainable version of BPA can be viewed as a meta-learning algorithm.
%, but it is \hl{transductive} (see ahead) and exploits the task items as a set, directly assessing the relative similarity relations between its items.

%% transductive
Subsequent works \cite{chen2018closer,dhillon2019baseline}  
% have questioned the benefits of meta-learning, advocating 
advocate 
%fine-tuning pre-trained networks, with 
using larger and more expressive backbones, employing \textit{transductive} inference, which fully exploits the data at inference, including unlabeled images.
% differentiation
BPA is transductive, but does not make assumptions on (nor needs to know) the number of classes (ways) or items per class (shots), as it executes a general probabilistic grouping action. 
%
%% attention based

Recently, \textit{attention} mechanisms were shown to be effective for FSC
%. We experiment with some leading works of this line 
\cite{ReNet,DeepEMD,ye2020few} %\simon{Did we? Or should we here?} % Relation Networks \cite{sung2018learning} replace the query-to-class calculations of ProtoNet~\cite{snell2017prototypical} with a learned network which is trained end-to-end and FEAT \cite{ye2020few} achieves excellent results by using set-set-functions, such as transformers \cite{vaswani2017attention}, to adapt the embedding to each classification task. 
% differentiation
%While BPA is clearly a self-attention method, it is parameterless and hence easier to employ in different networks, in particular with the ability to work without the need to retrain the entire network.
%% Sinkhorn
and a number of works \cite{ziko2020laplacian,huang2019few,hu2020leveraging,zhang2021sill,chen2021few} have adopted Sinkhorn~\cite{cuturi2013sinkhorn} as a parameterless unsupervised classifier that computes matchings between query embeddings and class centers. 
Sill-Net \cite{zhang2021sill} that augments training samples with illumination features and PTMap-SF \cite{chen2021few} that proposes DCT-based feature embedding, are both based on PTMap \cite{hu2020leveraging}. The state-of-the-art  PMF \cite{hu2022pushing},  proposed a 3 stage pipeline of pre-training on external data, meta-training with labelled tasks, and fine-tuning on unseen tasks.
% differentiation
%BPA uses \hl{Sinkhorn to solve an entirely different OT problem} - that of matching the set of features to itself, rather than against class representations. Nevertheless, 
BPA can be incorporated into these methods, immediately after their feature extraction stage.

\vspace{-7pt}
\paragraph{Unsupervised Image Clustering (UIC)}
% ~~~~~~~~~~~~~~~~~~~~~~~~~~~~~~~~~~~~~~~~
\hspace{-8pt}
%The study of clustering techniques has been an active area of research for longer than a century, with a plethora of versions of the problem, approaches for its solution or approximation, and applications in any field that involves data-analysis. 
%In computer vision, the classical approaches can be roughly categorized as centroid-based (like k-means) distribution-based (like Gaussian Mixture Models) or density-based (like mean-shift). 
% They have been used excessively with great success, for clustering visual data at all levels: pixels (e.g. segmentation), patches, feature vectors, entire images and point-clouds, among others.
%
% In recent years, with the general shift to learning-based methods, clustering has had a certain resurgence, especially as a significant enabler in the growing interest in unsupervised learning and tremendous advances it has recently seen. 
%
% We identify several recent lines of research on clustering, to which our method has potential relevance. 
%
%is a branch in unsupervised learning research, where the goal is to
is the task of grouping related images, without any label information, into representative clusters. Naturally, the ability to measure the similarities among samples is a crucial aspect of UIC. 

Recent methods have achieved tremendous progress in this task, towards closing the gap with supervised  counterparts.
%, by successfully exploiting the representation power of deep learning models. 
%This has lead to standardised evaluation protocols, on datasets ranging from Cifar-10~\cite{krizhevsky2009learning} to different subsets of ImageNet~\cite{deng2009imagenet}. 
The leading approaches %\cite{chang2017deep, chang2018deep, tao2020clustering, van2020scan, niu2021spice} 
directly learn to map images to labels, by constraining the training of an unsupervised classification model with different types of indirect loss functions. 
Prominent works in this area include DAC \cite{chang2017deep}, which recasts the clustering problem into a binary pairwise-classification framework and SCAN \cite{van2020scan} which builds on a pre-trained encoder that provides nearest-neighbor based constraints for training a classifier. 
% and different techniques, like exploiting meaningful distances between features extracted by a deep network, that can be used (e.g. by well designed augmentations, or by KNN-classifiers) to generate information that can be used directly by unsupervised clustering procedures or simple classifiers. 
The recent state-of-the-art SPICE \cite{niu2021spice}, is a pseudo-labeling based method, which divides the clustering network into a feature model for measuring the instance-level similarity and a clustering head for identifying the cluster-level discrepancy.

%UIC has recently gained popularity as a means for self-supervision in feature learning, showing excellent results on unsupervised image classification ~\cite{caron2018deep,caron2020unsupervised}.

%
%Clustering is a clear-case instance-specific problem, since most information is instance-relevant and unrelated directly to other training instances. The BPA transform can hence be used to upgrade the feature representation quality.
%\simon{Elaborate a little on methods/techniques? The above ones have some commented text.. But maybe replace IIC and SCAN with ones we compare to?}
%However, there remain several required further generalizations that are left for future work. On the other hand, in this work, we find the line of work on \textit{supervised} images clustering, an excellent example for the application of our transform. 
%We experiment on a synthetic unsupervised clustering task, with control on feature dimensionality and intra-cluster variance and plan to test on various clustering benchmarks in future work.

\vspace{-7pt}
\paragraph{Person Re-Identification (Re-ID)}
% ~~~~~~~~~~~~~~~~~~~~~~~~~~~~~~~~~~~~~~~~~~~
\hspace{-8pt}
is the task identifying a certain person (identity) between multiple detected pedestrian images, from different non-overlapping cameras. It is challenging due to the scale of the problem and large variation in pose, background and illumination.

% Re-ID is typically considered an instance retrieval problem and hence can be tackled using metric learning tools. The data is divided into a set of query images and a large set of gallery images, with the goal of finding a representation that minimizes the distance between query images and their corresponding gallery samples.

See \citet{ye2021deep} for an excellent comprehensive survey on the topic.
Among the most popular methods are OSNet~\cite{zhou2019osnet} that developed an efficient small-scale network with high performance and DropBlock (Top-DB-Net)~\cite{Top-DB-Net} which achieved state-of-the-art results by dropping a region block in the feature map for attentive learning.
The Re-ID task is typically larger scale - querying thousands of identities against a target of tens of thousands. Also, the data is much more real-world compared to the carefully curated FSC sets. 

% <<<<<<<<<<<<<<<<<<<<<<<<<

%-------------------------------------------------------------------
% method
%-------------------------------------------------------------------
\section{The BPA Transform}

\subsection{Derivation} \label{sec.transform.derivation}
%=======================================================================

Assume we are given a task instance which consists of an inference problem over a set of $n$ items $\{x_i\}_{i=1}^n$, where each of the items belongs to a space of input items $\Omega\subseteq\mathbb{R}^D$. %, for some fixed dimension $D$.
    %\item Assume that each item $x_i$ is described by some $d$-dimensional feature $v_i\in \mathbb{R}^d$
%
The inference task can be modeled as $f_\theta(\{x_i\}_{i=1}^n)$, using a learned function $f_\theta$, which acts on the set of input items and is parameterized by a set of parameters $\theta$.
Typically, such functions combine an initial feature extraction stage that is applied independently to each input item, with a subsequent stage of (separate or joint) processing of the feature vectors (see Fig. \ref{fig.network} {Left or Right}, respectively).% \simon{ref figure}

That is, the function $f_\theta$ takes the form $f_\theta(\{x_i\}_{i=1}^n)=G_{\psi}(\{F_{\phi}(x_i)\}_{i=1}^n)$, where $F_{\phi}$ is the feature extractor (or embedding network) and $G_{\psi}$ is the task inference function, parameterized by $\phi$ and $\psi$ respectively, where $\theta=\phi\cup\psi$. 

The feature embedding $F:\mathbb{R}^D\rightarrow\mathbb{R}^d$, usually in the form of a neural-network (with $d\ll D$), could be either pre-trained, or trained in the context of the task function $f$, along with the inference function $G$.

For an input $\{x_i\}_{i=1}^n$, let us define the set of embedded features $\{v_i\}_{i=1}^n=\{F(x_i)\}_{i=1}^n$. In the following, we consider these sets of input vectors and features as real-valued row-stacked matrices $\mathcal{X}\in \mathbb{R}^{n\times D}$ and $\mathcal{V}\in\mathbb{R}^{n\times d}$.

We suggest a novel re-embedding of the feature set $\mathcal{V}$, using a transform, that we denote by $T$, in order to obtain a new set of features $\mathcal{W}=T(\mathcal{V})$, where $\mathcal{W}\in\mathbb{R}^{n\times n}$.
The new feature set $\mathcal{W}$ has an explicit probabilistic interpretation, which is specifically suited for tasks related to classification, matching or grouping of items in the input set $\mathcal{X}$.
In particular, $\mathcal{W}$ will be a symmetric, doubly-stochastic matrix (non-negative, with rows and columns that sum to $1$), where the entry $w_{ij}$ (for $i\neq j$) encodes the belief that items $x_i$ and $x_j$ belong to the same class or cluster. 

The proposed transform \mbox{$T:\mathbb{R}^{n\times d}\rightarrow \mathbb{R}^{n\times n}$} (see Fig. \ref{fig.BPA_illustration}) acts on the original feature set $\mathcal{V}$ as follows. It begins by computing the squared Euclidean pairwise distances matrix $\mathcal{D}$, namely, $d_{ij}=||v_i-v_j||^2$, which can be computed efficiently as $d_{ij}=2(1-cos(v_i,v_j))=2(1-v_i\cdot v_j^T)$, when the rows of $\mathcal{V}$ are unit normalized. Or in a compact form, $\mathcal{D}=2(\textbf{1}-\mathcal{S})$, where
$\textbf{1}$ is the all ones $n\times n$ matrix and 
$\mathcal{S}=\mathcal{V}\cdot\mathcal{V}^T$ is the cosine affinity matrix of $\mathcal{V}$. 

$\mathcal{W}$ will be computed as the optimal transport (OT) plan matrix between the $n$-dimensional all-ones vector $\textbf{1}_n$ and itself, under the self cost matrix $\mathcal{D}_\infty$, which is the distance matrix $\mathcal{D}$ with a very (infinitely) large scalar replacing each of the entries on its diagonal (which were all zero), that enforces the affinities of each feature to distribute among the others.
Explicitly, let $\mathcal{D}_\infty=\mathcal{D}+\alpha I$, where $\alpha$ is a very (infinitely) large constant and $I$ is the $n\times n$ identity matrix.

$\mathcal{W}$ is defined to be the doubly-stochastic matrix that is the minimizer of the functional
    \begin{equation} \label{eq.fractional_matching}
      \mathcal{W}=
      \argminA_{\mathcal{W}\in B_n}\:\langle \mathcal{D}_\infty,\mathcal{W}\rangle
    \end{equation}
    where $B_n$ is the set (known as the Birkhoff polytope) of $n\times n$ doubly-stochastic matrices and $\langle\cdot,\cdot\rangle$ stands for the Frobenius (standard) dot-product.

This objective can be minimized using simplex or interior point methods with complexity $\Theta(n^3\log n)$. In practice, we use the highly efficient Sinkhorn-Knopp method \cite{cuturi2013sinkhorn}, which is an iterative scheme that optimizes an entropy-regularized version of the problem, where each iteration takes $\Theta(n^2)$. Namely:
    \begin{equation} \label{eq.entropy_min}
        \mathcal{W}=\argminA_{\mathcal{W}\in B_n}\:\langle \mathcal{D}_\infty,\mathcal{W}\rangle-\frac{1}{\lambda}h(\mathcal{W})
    \end{equation} \label{eq:sinkhorn_objective}
where $h(\mathcal{W})=-\sum_{i,j} w_{ij} \log(w_{ij})$ is the Shannon entropy of $\mathcal{W}$ and $\lambda$ is the entropy regularization parameter. 

The \textit{transport-plan} matrix $\mathcal{W}$ that is the minimizer of \cref{eq.entropy_min} will become the result of our transform, after 
%rescaling the dynamic range of the matrix (dividing by its maximal value) and 
'restoring' perfect affinities on the diagonal (replacing the diagonal entries from $0$s to $1$s) by $\mathcal{W}=\mathcal{W}+I$, where $I$ is the $n\times n$ identity matrix. Our final set of features is $T(\mathcal{V})=\mathcal{W}$ and each of its rows is the re-embedding of each of the corresponding features (rows) in $\mathcal{V}$. 
The BPA transform is given in \cref{algo:BPA} in the appendix, in PyTorch-style pseudo-code.
Note that $\mathcal{W}$ is symmetric as a result of the symmetry of $\mathcal{D}$ and its own double-stochasticity. We next explain its probabilistic interpretation.

\subsection{Probabilistic interpretation} \label{sec.transform.interpretation}
%========================================================================

The optimization problem in \cref{eq.fractional_matching} can be written more explicitly as follows:
        \begin{equation}
        \begin{aligned}
        \min_{\mathcal{W}} \; \langle \mathcal{D}_\infty,\mathcal{W}\rangle \quad \quad
        \textrm{s.t.} \quad \quad & \mathcal{W}\cdot \textbf{1}_n=\mathcal{W}^T\cdot \textbf{1}_n= \textbf{1}_n
        \end{aligned} \label{eq.opt_D_inf}
        \end{equation}

 which can be seen to be the same as:
        \begin{equation}
        \begin{aligned}
        \min_{\mathcal{W}} \; \langle \mathcal{D},\mathcal{W}\rangle 
        \quad \quad
        \textrm{s.t.} \quad \quad & \mathcal{W}\cdot \textbf{1}_n=\mathcal{W}^T\cdot \textbf{1}_n= \textbf{1}_n \\
        & w_{ii}=0 \quad\text{for}\quad i = 1,\dots n
        \end{aligned} \label{eq.opt_fractional_matching}
        \end{equation}
        since the use of the infinite weights on the diagonal of $\mathcal{D}_\infty$ is equivalent to using the original $\mathcal{D}$ with a constraint of zeros along the diagonal of $\mathcal{W}$.

\begin{figure}[t!]
\centering
\vspace{-9pt}
\includegraphics[width=0.88\columnwidth]
{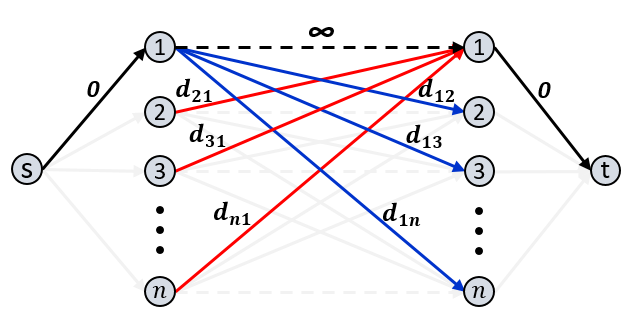}
\vspace{-10pt} 
\caption{
{\small \selectfont
{\textbf{The min-cost max-flow perspective}: Costs are shown.}}}
\label{fig.min-cost max-flow}  \vspace{-12pt} 
\end{figure}

The optimization problem in \cref{eq.opt_fractional_matching} is in fact a fractional matching instance between the set of $n$ original features and itself. It can be posed as a bipartite-graph min-cost max-flow instance (The problem of finding a min cost flow out of all max-flow solutions), as depicted in Fig.~\ref{fig.min-cost max-flow}. The graph has $n$ nodes on each side, representing the original features $\{v_i\}_{i=1}^n$ (the rows of $\mathcal{V}$). Across the two sides, the cost of the edge $(v_i,v_j)$ is the distance $d_{ij}$ and the edges of the type $(v_i,v_i)$ have a cost of infinity (or can simply be removed). Each `left' node is connected to a 'source' node S by an edge of cost $0$ and similarly each 'right' node is connected to a `target' (sink) node T. All edges in the graph have a capacity of $1$ and the goal is to find an optimal fractional self matching, by finding a min-cost max-flow from source to sink. Note that the max-flow can easily be seen to be $n$, but a min-cost flow is sought among  max-flows.

%% -------------------- min-cost max-flow figure ----------\
%% ------------------------- figure ------------------/

In this set-to-itself matching view, each vector $v_i$ is fractionally matched to the set of all other vectors $\mathcal{V}-\{v_i\}$  based on the pairwise distances, but importantly taking into account the fractional matches of the rest of the vectors in order to satisfy the double-stochasticity constraint. The construction constrains the max flow to have a total outgoing flow of 1 from each `left' node and a total incoming flow of 1 to each `right' node.
Therefore, the $i$th transformed feature $w_i$ ($i$th row of $\mathcal{W}$) is a \textit{distribution} (non-negative entries, summing to $1$), where $w_{ii}=0$ and  $w_{ij}$ is the relative belief that features $i$ and $j$ belong to the same `class'.

\subsection{Properties} \label{sec.transform.properties}
%========================================================================

We can now point out some important properties of the proposed embedding, given by the rows of the matrix $\mathcal{W}$. Some of these properties can be observed in the toy 3-class MNIST digit example, illustrated in Fig.~\ref{fig.BPA_illustration}.

\begin{figure}[t]
\centering
\vspace{-3pt}
\setlength{\tabcolsep}{6pt} % for the horizontal padding
\hspace{-1pt}
\includegraphics[width=1.0\columnwidth]
{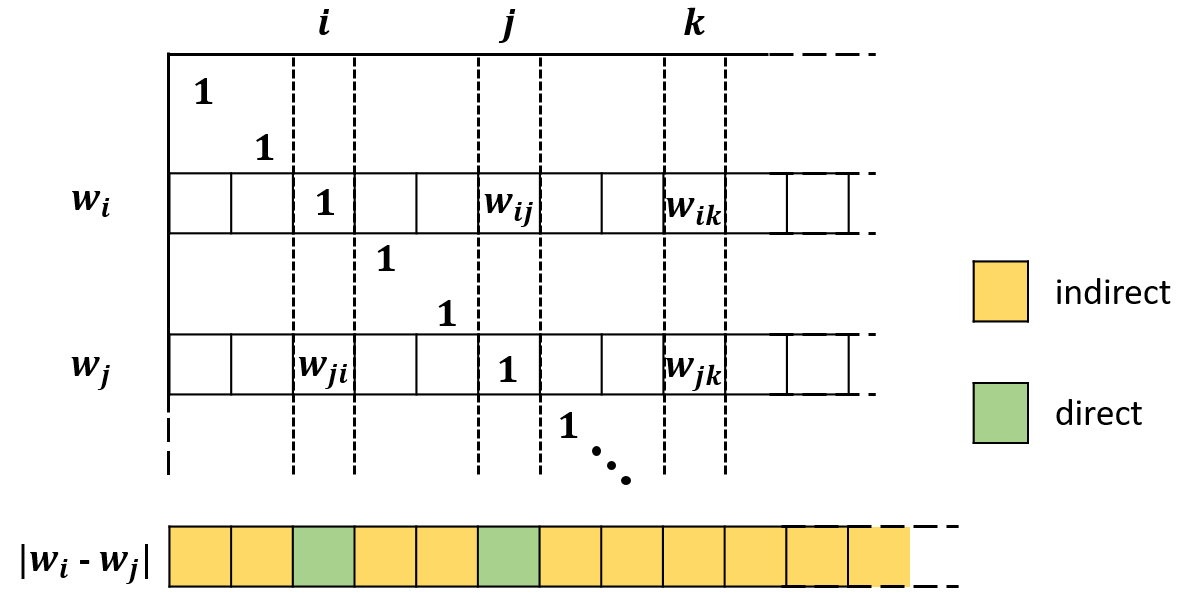} \vspace{-18pt}
\caption{
{\small \selectfont
{\textbf{The (symmetric) embedding matrix $\mathcal{W}$} and the absolute difference between its $i$th and $j$th rows.}}}
%: We examine the vector $|w_i-w_j|$: (i) Its $i$th and $j$th coordinates equal $|1-w_{ij}|=|1-w_{ji}|$, giving the \textit{direct} similarity between the original features, since this amount (in green) is greater when $w_{ij}$ and $w_{ji}$ (the mutual beliefs) are high (closer to 1). ; (ii) Its $k$th coordinate (for any $k\notin\{i,j\}$) gives $|w_{ik}-w_{jk}|$ which is an \textit{indirect} (third-party) comparison between the original features through the $k$th feature. Similarity (in yellow) is stronger when features $i$ and $j$ have a similar belief regarding feature $k$, \textit{i.e.} $w_{ik}$ and $w_{jk}$ are close.}}}
\label{fig.embedded_differences}  \vspace{-14pt} 
\end{figure}

\vspace{3pt}\noindent \textbf{Interpretability of distances in the embedded space}:
%=============================================================
%\vspace{3pt}\noindent \textbf{Direct and Indirect similarity encoding}:
An important property of our embedding is that each embedded feature encodes its distribution of affinities to all other features. In particular, the comparison of embedded vectors $w_i$ and $w_j$ (of items $i$ and $j$ in a set) includes both \textit{direct} and \textit{indirect} information about the similarity between the features. Refer to \cref{fig.embedded_differences} for a detailed explanation of this property. 
If we look at the different coordinates $k$ of the absolute difference vector $a=|w_i-w_j|$, BPA captures
(i) \textit{direct affinity}: For $k$ which is either $i$ or $j$, it holds that \mbox{$a_k=1-w_{ij}=1-w_{ji}$} 
\footnote{Note: (i) $w_{ii}=w_{jj}=1$ ; (ii) $w_{ij}=w_{ji}$ from  symmetry of $\mathcal{W}$ ; (iii) all elements of $\mathcal{W}$ are $\leq 1$ hence the $|\cdot|$ can be dropped ;}.
This amount measures how high (close to $1$) is the mutual belief of features $i$ and $j$ about one another.
(ii) \textit{indirect (3rd-party) affinity}: For $k\notin\{i,j\}$, we have $a_k=|w_{ik}-w_{jk}|$, which is a comparison of the beliefs of features $i$ and $j$ regarding the (third-party) feature $k$.
The double-stochasticity of the transformed feature-set ensures that the compared vectors are similarly scaled (as distributions, plus 1 on the diagonal) and the symmetry further enforces the equal relative affinity between pairs.

%% ------------------------- figure W-matrix ----------\
%% ------------------------- figure ------------------/

As an example, observe the output features 4 and 5 in Fig.~\ref{fig.BPA_illustration}, that re-embed the 'green' features of the digit '7' images. As desired, these embedding are close in the target 7D space. The closeness is driven by both their closeness in the original space (coordinates 4 and 5) as well as the agreement on specific large differences from other images. This property is responsible for better separation between classes in the target domain, which leads to improved performance on tasks like classification, clustering or retrieval.

% The non-parametric nature gives BPA an advantage over other set-to-set methods such as transformers in that its output is interpretable (e.g. by  inspecting the transport-plan matrix $\mathcal{W}$), with a clear probabilistic characterization of the found relations (Sec. \ref{sec.transform.interpretation}). 
%As an example, observe the output features 1, 3 and 6 in Fig.~\ref{fig.BPA_illustration}, that re-embed the 'red' features of the digit '4' images. Each of these vectors is a 7-dim distribution of relative similarities, that share a similar distribution agreeing about the high values at indices 1, 3 and 6 and low values for the rest.

    % \begin{itemize}
    % \setlength{\itemindent}{-0.8em}
    % \\ \vspace{-2pt}
    % \item \textbf{direct similarity}:
    %     For $k$ which is either $i$ or $j$, it holds that $a_k=1-w_{ij}=1-w_{ji}$ \footnote{Note: (i) $w_{ii}=w_{jj}=1$ ; (ii) $w_{ij}=w_{ji}$ from the symmetry of $\mathcal{W}$ ; (iii) all elements of $\mathcal{W}$ are $\leq 1$ and hence the $|\cdot|$ can be dropped ;}. This amount measures how high (\ie close to $1$) is the mutual belief of features $i$ and $j$ about one another.
    %     \\ \vspace{-14pt}
    % \item \textbf{indirect (3rd-party) similarity}: For $k\notin\{i,j\}$, we have            $a_k=|w_{ik}-w_{jk}|$, which is a comparison of the beliefs of features $i$ and     $j$ regarding the (third-party) feature $k$.
    %     \\ \vspace{-17pt}
    % \end{itemize}

\vspace{3pt}\noindent \textbf{Parameterless-ness}, \textbf{Differentiability} and \textbf{Equivariance}:
%=============================================================
These three properties are inherited from the Sinkhorn OT solver. The transform is parameterless, giving it the flexibility to be used in other pipelines, directly over different kinds of embeddings, without the harsh requirement of retraining the entire pipeline. Retraining is certainly possible, and beneficial in many situations, but not mandatory, as our experiments work quite well without it. 
Also, due to the differentiability of the Sinkhorn algorithm \cite{cuturi2013sinkhorn}, back-propagating through BPA can be done naturally, hence it is possible to (re-)train the hosting network to adapt to BPA, if desired.
The embedding works coherently with respect to any change of order of the input items (features). This can be shown by construction, since  min-cost max-flow solvers as well as the Sinkhorn OT solver are equivariant with respect to permutations of their inputs.
%that if $\mathcal{W}=T(\mathcal{V})$ and $\mathcal{W}^\pi=T(\pi(\mathcal{V}))$ is the transform applied on a permutation $\pi$ of the rows of $\mathcal{V}$, then for any $i$ it holds that $\|w_i-w_j\|_p=\|w^\pi_{\pi(i)}-w^\pi_{\pi(j)}\|_p$.
% pair $i$ and $j$ and for every $\ell_p$ norm it holds that $\|w_i-w_j\|_p=\|w^\pi_{\pi(i)}-w^\pi_{\pi(j)}\|_p$.

\vspace{1pt}\noindent \noindent\textbf{Usage flexibility}:
%=============================================================
Recall that BPA is applied on sets of features, typically computed by some embedding network and its output features are passed to downstream network components. Since BPA is \textit{parameterless}, it can be simply inserted to any trained hosting network and since it is \textit{differentiable}, it is possible to train the hosting network with BPA inside it. 
We therefore denote by \text{\BPAp} the basic drop-in usage of BPA, inserted into a \blue{\textit{pretrained}} network. This is the easiest and most flexible way to use BPA, nevertheless showing consistent benefits in the different tested applications. 
We denote by \text{\BPAt} the usage where the hosting network is \red{\textit{trained}} with BPA within. It allows to adapt the hosting network's parameters to the presence of the transform, with the potential of further improving performance. 

\vspace{1pt}\noindent \noindent\textbf{Transductive or Inductive}:
%=============================================================
Note that BPA is a \textit{transductive} method in the sense that it needs to jointly process the data, but in doing so, unlike many transductive methods, it does not make any limiting assumptions about the input structure, such as knowing the number of classes, or items per class. In any case, we consider the \text{\BPAp} and \text{\BPAt} variants to be transductive, regardless of the nature of the hosting network.
Nevertheless, being transductive is possibly restrictive for certain tasks, for which test-time inputs might be received one-by-one. Therefore, we suggest a third usage type, \text{\BPAi}, where the hosting network is trained with BPA inside (just like in \text{\BPAt}), but BPA is not applied at inference (simply not inserted), hence the hosting network remains \cyan{\textit{inductive}} if it was so in the first place.

\vspace{1pt}\noindent \textbf{Dimensionality}:
%=============================================================
BPA has the unique property that the dimension of its embedded feature depends on (equals) the number of features in the set. Given a batch of $n$ $d$-dimensional features $V\in \mathbb{R}^{n\times d}$, it outputs a batch of $n$ $n$-dimensional features $W={\BPAreg}(V)\in \mathbb{R}^{n\times n}$. On one hand, this is a desired property, since it is natural that the feature dimensionality (and capacity) depends on the complexity of the task, which typically grows with the number of features (Think of the inter-relations which are more complex to model). On the other hand, it might impose a problem in situations at which the downstream calculation that follows expects a specific feature dimension, for example with a pre-trained non-convolutional layer.

In order to make BPA usable in such cases, we propose an attention-like variant, \text{\textbf{BPA\_Attn}}, in which the normalized BPA matrix is used to balance the input features without changing their dimension, by simple multiplication, i.e. 
$\text{\textbf{BPA\_Attn}}(V)=\BPAreg(V)\cdot V$. This variant allows to maintain the original feature dimension $d$, or even a smaller dimension if desired, by applying dimension reduction on the original set of features prior to applying $\text{\textbf{BPA\_Attn}}$.

%Of particular interest are the options of applying \textbf{BPA\_Attn} on either:
%(i) the original features, hence keeping the original dimension, allowing the most simple insertion of BPA into an existing pipeline; (ii) features that were down-sampled by dimensionality reduction. 
% Nevertheless, in many situations the downstream computation has the flexibility to work with varying input dimensions, allowing to work with the vanilla \textbf{BPA}. 
In Table~\ref{tab:dimensions_5w1s}, we examine few-shot classification accuracy on \textit{MiniImagenet} \cite{vinyals2016matching} with downstream classification by ProtoNet \cite{snell2017prototypical}. Each classification instance consists of 80 images, encoded to 640-dimensional features by a pre-trained resnet-12 network. ProtoNet works on either: (i) the original feature set $V$ (ii) its dimension reduced versions, calculated by either PCA or Spectral-Clustering (SC) (iii) vanilla \textbf{BPA} (iv) \textbf{BPA\_Attn} on original or reduced features.
As can be observed, the best accuracies are achieved by vanilla \textbf{BPA}, but the attention provided by BPA is able to stabilize performance across the entire range of dimensions. 

%Also, in most benchmarks the {instance set sizes are fixed}, allowing for a single setting of sizes to work throughout.

\newcommand{\wayshot}[2]{\textbf{\textbf{\textit{#1}}-way \textbf{\textit{#2}}-shot}}

\begin{table}[t]
    \vspace{-6pt}
    \centering
    \caption{\textbf{Feature-dimension control strategies}: Accuracy on $\wayshot{5}{1}$ \textit{MiniImagenet}. * marks the dimension of original 640d pre-trained resnet-12 features. \# marks the size of a batch that includes a single 5-way 1-shot 15-query task ($80=5\cdot(1+15)$), which is the output dimension of vanilla \textbf{BPA}. Best and second best results, per dimension, are in \textbf{Bold} and \textit{\textbf{italics}}.}
    \vspace{5pt}
    \fontsize{8.1}{8.1} \selectfont
    \setlength{\tabcolsep}{0.6em} % for the horizontal padding
    \renewcommand{\arraystretch}{1.3}
\begin{tabular}{|l |  c c c c c c|}
\hline
% &   \multicolumn{6}{c|}{\textbf{\textit{dimension}}} \\
% \hline
$\:$\textbf{\textit{input to ProtoNet} \slash  \textit{ dim.}} &       5  &  10  &  20  &  40   &  80$^\#$  &  640$^{*}$ \\
\hline 
$V$  (orignal)     & -      &   -    &   -  &  -     &    -   & 64.6\\
PCA($V$)               &{66.2}&{65.7}&{64.4}&{64.1}&{64.3}&-\\

SC($V$)          &{66.8}&{58.2}&{46.2}&{38.3}&{25.5}&-\\
\textbf{\BPAps}($V$)               & -      &   -    &     -  &  -     &\textit{\textbf{71.2}}&  -  \\
\textbf{\BPAts}($V$)               & -      &   -    &     -  &  -     &\textbf{72.1}&  -  \\
\hdashline
\textbf{\BPAps\_Attn}($V$) & -      &   -    &     -  &  -     &    -   & \textbf{\textit{69.1}}\\
\textbf{\BPAts\_Attn}($V$) & -      &   -    &     -  &  -     &    -   & \textbf{70.0}\\
\textbf{\BPAps\_Attn}(SC($V$)) &{\textbf{69.1}}&{\textbf{69.1}}&{\textbf{68.1}}&{\textbf{68.5}}&69.2&-\\
\hdashline
\textbf{\BPAps\_Attn}(PCA($V$))      &\textit{\textbf{67.1}}&\textit{\textbf{67.8}}&\textit{\textbf{67.5}}&\textit{\textbf{67.6}}&{67.8}&-\\
\hline
\end{tabular}\vspace{-12pt}
    \label{tab:dimensions_5w1s} 
\end{table}

\newcommand{\mc}{\multicolumn{2}{|c|}}

%=============================================================
\newcommand{\kkk}[1]{({\footnotesize #1})}
\newcommand{\bbb}[1]{({\fontsize{8.1}{8.1}\selectfont\color{blue}#1})}
\newcommand{\rrr}[1]{({\fontsize{8.1}{8.1}\selectfont\color{red} #1})}
\newcommand{\ccc}[1]{({\fontsize{8.1}{8.1}\selectfont\color{cyan} #1})}
\newcommand{\cmark}{\hspace{-10pt}\ding{51}\hspace{-6pt}}%
\newcommand{\xmark}{\hspace{-10pt}\ding{55}\hspace{-6pt}}%
%=============================================================
%\newcommand{\wayshot}[2]{\textbf{\textbf{\textit{#1}}-way \textbf{\textit{#2}}-shot}}

%%%%%%%%%%%%%%%%%%%%%%%%%%%%%%%
%%%%%%%%%% I-Net %%%%%%%
%%%%%%%%%%%%%%%%%%%%%%%%%%%%%%%
%===============================================
%----------------------------- MiniImagenet -------------
\begin{table}[t]
    \centering\vspace{-12pt}
    \caption{
    {%\fontsize{8.5}{8.5} \selectfont
    \textbf{Few-Shot Classification (FSC)} accuracy on \textit{MiniImagenet}. Results are ordered by backbone (resnet-12, wrn-28-10, ViT small/base), each listing baseline methods and  BPA variants. BPA improvements (colored percentages) are in comparison with each respective baseline hosting method (obtained by division). 
    \textbf{Bold} and \textit{\textbf{italics}} highlight best and second best results per backbone.
    \textbf{\textit{T/I}} denotes transductive/inductive methods. 
        % (*) = from \cite{chen2018closer} ;
        (\&) from \citet{ziko2020laplacian}; 
        (\$) from original paper; 
        (\#) our implementation;
        }}
    \vspace{3pt}    
    \centering 
    \fontsize{8.1}{8.1} \selectfont
    \setlength{\tabcolsep}{0.5em} % for the horizontal padding
    \renewcommand{\arraystretch}{1.2}
    \begin{tabular}{lcccc}%{|l|l|c|c|}
    \hline    
    {\textbf{method}}    &  {\textbf{\textit{T/I}}}    &  {\textbf{network}}  &  \wayshot{5}{1}  &  \wayshot{5}{5}   \\\hline 
%     MAML(*)        \cite{finn2017model}& \textit{I}  & conv-4    &  46.47  &  62.71  \\ 
%     RelationNet(*) \cite{sung2018learning}\!\!\!\!\!\!\!\!& \textit{I}  & conv-4    &  49.31  &  66.60  \\ 
%     ProtoNet(\#) \cite{snell2017prototypical} & \textit{I}  & conv-4    &  49.10  &  66.79  \\ 
%     FEAT(\$) \cite{ye2020few}   & \textit{I}  & conv-4 &  \textbf{55.15}  &  \textbf{71.61}  \\ 
% \hdashline
%     ProtoNet-\BPAp  & \textit{T}  & conv-4    &  \textbf{\textit{54.01}} \bbb{ +10.2\%}  &  69.39 \bbb{+3.9\%}  \\   
%     ProtoNet-\BPAts  & \textit{T}  & conv-4    &  53.70 \rrr{+9.3\%}  &  \textbf{\textit{70.40}} \rrr{+5.4\%}  \\    
% \hline
    ProtoNet(\#) & \hspace{-2pt}\textit{I}  & ResNet &  62.39  &  80.33  \\ 
    DeepEMD(\$)  & \hspace{-2pt}\textit{I}  & ResNet &  65.91  &  82.41  \\
    FEAT(\$)     & \hspace{-2pt}\textit{I}  & ResNet &  66.78  &  82.05  \\
    RENet(\$)    & \hspace{-2pt}\textit{I}  & ResNet &  67.60  &  82.58  \\
    % PTMap(\#) \cite{hu2020leveraging}  & \textit{T}  & ResNet &  76.90  &  85.20  \\ 
\hdashline
    ProtoNet-\BPAps  & \hspace{-2pt}\textit{T}  & ResNet &  67.34  \bbb{+7.9\%}  & 81.84 \bbb{+1.6\%} \\   
    ProtoNet-\BPAis  & \hspace{-2pt}\textit{I}  & ResNet &  64.36 \ccc{+3.1\%}  &  81.82 \ccc{+1.8\%}  \\    
    ProtoNet-\BPAts  & \hspace{-2pt}\textit{T}  & ResNet &  67.90 \rrr{+8.8\%}  &  83.09 \rrr{+3.2\%}  \\
    % PTMap\BPAps     & \textit{T}  & ResNet &  \textbf{78.35} \bbb{+1.9\%}  &  \textbf{86.01} \bbb{+1.0\%}  \\   
    % PTMap\BPAts     & \textit{T}  & ResNet &  \textbf{\textit{77.30}} \rrr{+0.5\%}  &  \textbf{\textit{85.49}} \rrr{+0.3\%}  \\    
    \hline
    ProtoNet(\&) & \hspace{-2pt}\textit{I}  & WRN &  62.60  &  79.97  \\ 
    PTMap(\$)   & \hspace{-2pt}\textit{T}  & WRN &  82.92  &  88.80  \\ 
    SillNet(\$) & \hspace{-2pt}\textit{T}  & WRN &  82.99  &  89.14  \\ 
    PTMap-SF(\$)  & \hspace{-2pt}\textit{T}  & WRN &  \textbf{\textit{84.81}}  &  \textbf{\textit{90.62}}  \\ 
% \hdashline
%     PTMap-cosine     & \textit{T}  & WRN &  74.60 \kkk{-10.0\%} &  84.68 \kkk{-4.6\%} \\   
%     PTMap-softmax    & \textit{T}  & WRN &  80.08 \kkk{-3.4\%} &  83.83 \kkk{-5.6\%} \\
\hdashline
    PTMap-\BPAps     & \hspace{-2pt}\textit{T}  & WRN &  83.19 \bbb{+0.3\%} &  89.56 \bbb{+0.9\%} \\   
    PTMap-\BPAts     & \hspace{-2pt}\textit{T}  & WRN &  84.18 \rrr{+1.5\%} &  90.51 \rrr{+1.9\%} \\
    SillNet-\BPAps  & \hspace{-2pt}\textit{T}  & WRN &  83.35 \bbb{+0.4\%} &  89.65 \bbb{+0.6\%} \\
    PTMap-SF-\BPAps  & \hspace{-2pt}\textit{T}  & WRN &  \textbf{85.59} \bbb{+0.9\%}  &  \textbf{91.34}  \bbb{+0.8\%} \\
\hline  
% ViT-s
PMF(\$)   & \hspace{-2pt}\textit{I}  & ViT-s &  \textit{93.10} &  \textbf{98.00}  \\
\hdashline
PMF-\BPAps  & \hspace{-2pt}\textit{T}  & ViT-s &  94.49 \bbb{+1.4\%} &  97.68 \bbb{-0.3\%} \\
PMF-\BPAis  & \hspace{-2pt}\textit{I}  & ViT-s &  92.70 \ccc{-0.4\%} &  \textbf{98.00} \ccc{+0.0\%} \\
    PMF-\BPAps  & \hspace{-2pt}\textit{T}  & ViT-s &  \textbf{95.30} \rrr{+2.3\%}  &  \textit{\textbf{97.90}}  \rrr{-0.1\%} \\
\hline 
% ViT-b
PMF(\$)    & \hspace{-2pt}\textit{I}  & ViT-b &  \textit{\textbf{95.30}} &  98.40  \\
\hdashline
PMF-\BPAps  & \hspace{-2pt}\textit{T}  & ViT-b &  95.90 \bbb{+0.6\%} &  98.30 \bbb{-0.1\%} \\
PMF-\BPAis  & \hspace{-2pt}\textit{I}  & ViT-b &  95.20 \ccc{-0.1\%} &  \textbf{98.70} \ccc{+0.3\%} \\
    PMF-\BPAts  & \hspace{-2pt}\textit{T}  & ViT-b &  \textbf{96.3} \rrr{+1.0\%}  &  \textit{\textbf{98.5}}  \rrr{+0.1\%} \\
\hline
    \end{tabular}
    \label{tab:results_fsc_MiniImagenet} \vspace{-12pt}
\end{table} 
%-------------------------------------------------------------
%%%%%%%%%%%%%%%%%%%%%%%%%%%%%%%
%%%%%%%%%% end I-Net %%%%%%%
%%%%%%%%%%%%%%%%%%%%%%%%%%%%%%%

\vspace{2pt}\noindent 
\noindent\textbf{Hyper-parameters and ablations}:
%=============================================================
BPA has two hyper-parameters that were chosen through cross-validation and  kept fixed for each application over all datasets. The number of Sinkhorn iterations for computing the optimal transport plan was fixed to 5 and entropy regularization parameter $\lambda$  (Eq. \eqref{eq:sinkhorn_objective}) was set to 0.1 for UIC and FSC and to 0.25 for ReID. 
% \vspace{2pt}\noindent 
% \noindent\textbf{Further ablations}: 
% %=============================================================
In Appendix \ref{sec:ablations} we ablate these hyper-parameters as well as the scalability of BPA in terms of set-input size (Fig. \ref{fig.scaling}) on few-shot-classification, and in Appendix \ref{sec:case_study}, we study its robustness to noise and feature dimensionality (Fig. \ref{fig.synthetic_exp}) by a controlled synthetic clustering experiment.

%We also look into the effect of the features transform, by comparing the features before and after the transform, through a t-SNE embedding on a 10-way 10-shot mini-Imagenet instance. The  middle panel (after transform) shows improved grouping of the clustered images. The right panel compares the PDFs of inter and intra class pairwise distances in the embedding space (before and after the transform), showing an improved separation between the distributions after the application of the transform.

% %% ------------------------- figure -------------------
% \begin{figure*}
%   \centering
% \begin{tabular}{ccc}
% \includegraphics[height=3.9cm]{figures/4_features.pdf}&
% \includegraphics[height=3.9cm]{figures/4_OT.pdf}&
% \includegraphics[height=3.9cm]{figures/4_pdfs.pdf}\\
% original embedding & transformed embedding & inter vs intra distances
% \end{tabular}
%     \caption{Caption for the whole figure}
%     \label{fig.point_dist} % I can do without the label too
% \end{figure*}
% %% ------------------------- figure ------------------- % <<<<<<<<<<<<<<<<<<<<<<<<<

%-------------------------------------------------------------------
% Implementation details
%-------------------------------------------------------------------
% \section{Implementation details}

% \input{implementation_details} % <<<<<<<<<<<<<<<<<<<<<<<<<

%-------------------------------------------------------------------
% experimental results
%-------------------------------------------------------------------
\section{Results}

In this section, we experiment with BPA on three applications: Few-Shot Classification (Sec. \ref{sec.res.fsc}), Unsupervised Image Clustering (Sec. \ref{sec.res.clustering}) and Person Re-Identification (Sec. \ref{sec.res.reid}).
In each, we achieve state-of-the-art results, by merely using current state-of-the-art methods as hosting networks of the BPA transform. Perhaps more importantly, we demonstrate the flexibility and simplicity of applying BPA in these setups, with improvements in the entire range of testing, including different hosting methods, different feature embeddings of different complexity backbones and whether retraining the hosting network or just dropping-in BPA and performing standard inference.
To show the simplicity of inserting BPA into hosting algorithms, we provide pseudocodes for each of the experiments in Appendix \ref{sec:insertions}.

%-----------------------------  Cifar -------------
\begin{table}[t!]
    \centering\vspace{-12pt}
    \caption{
    {
    \textbf{\hspace{-4pt}Few-Shot Classification (FSC)} accuracy on \textit{CIFAR-FS}.
        % (\$) = from the method's paper itself ; 
        }}   
        \vspace{3pt} 
    \fontsize{8.1}{8.1} \selectfont
    \setlength{\tabcolsep}{0.5em} % for the horizontal padding
    \renewcommand{\arraystretch}{1.2}
    \begin{tabular}{lcccc}%{|l|l|c|c|}
    \hline    
    {\textbf{method}}    &  \hspace{-3pt}{\textbf{\textit{T/I}}}    &  {\textbf{network}}  &  \wayshot{5}{1}  &  \wayshot{5}{5}   \\
    \hline
    PTMap(\$)  & \textit{T}  & WRN &  87.69  &  90.68  \\ 
    SillNet(\$)  & \textit{T}  & WRN &  87.73  &  91.09  \\ 
    PTMap-SF(\$)   & \textit{T}  & WRN &  \textbf{\it{89.39}}  &  \textbf{\it{92.08}}  \\ 
\hdashline
    PTMap-\BPAps     & \textit{T}  & WRN &  87.37  \bbb{-0.4\%} &  91.12  \bbb{+0.5\%} \\   
%    PTMap\BPAt     & \textit{T}  & WRN &  ++++ \rrr{+1.5\%} &  ++++ \rrr{+1.9\%} \\
    SillNet-\BPAps  & \textit{T}  & WRN &  87.30  \bbb{-0.5\%} &  91.40 \bbb{+0.3\%} \\
    PTMap-SF-\BPAps  & \textit{T}  & WRN &  \textbf{89.94} \bbb{+0.6\%}  &  \textbf{92.83} \bbb{+0.8\%} \\
\hline  
% ViT-s
PMF(\$)  & \textit{I}  & ViT-s &  81.1 &  92.5  \\
\hdashline
PMF-\BPAps  & \textit{T}  & ViT-s &  84.7 \bbb{+4.4\%} &  92.8 \bbb{+0.3\%} \\
PMF-\BPAis  & \textit{I}  & ViT-s &  \it{\textbf{84.80}} \ccc{+4.5\%} &  \it{\textbf{93.40}} \ccc{+0.9\%} \\
PMF-\BPAts  & \textit{T}  & ViT-s &  \textbf{88.90} \rrr{+9.6\%}  &  \textbf{93.80}  \rrr{+1.4\%} \\
\hline 
% ViT-b
PMF(\$)   & \textit{I}  & ViT-b &  84.30 &  92.20  \\
\hdashline
PMF-\BPAps  & \textit{T}  & ViT-b &  88.2 \bbb{+4.6\%} &  94 \bbb{+1.9\%} \\
PMF-\BPAis  & \textit{I}  & ViT-b &  \it{\textbf{87.10}} \ccc{+3.3\%} &  \it{\textbf{94.70}} \ccc{+2.7\%} \\
    PMF-\BPAts  & \textit{T}  & ViT-b &  \textbf{91.00} \rrr{+7.9\%}  &  \textbf{95.00}  \rrr{+3.0\%} \\
\hline
    \end{tabular}  \vspace{-12pt}       
    \label{tab:results_fsc_Cifar}
\end{table} 
%-------------------------------------------------------------
%%%%%%%%%%%%%%%%%%%%%%%%%%%%%%%
%%%%%%%%%% end CIFAR %%%%%%%
%%%%%%%%%%%%%%%%%%%%%%%%%%%%%%%

\subsection{Few-Shot Classification (FSC)} \label{sec.res.fsc}
Our main experiment is a comprehensive evaluation on the standard few-shot classification benchmarks \emph{MiniImagenet} \cite{vinyals2016matching} and \emph{CIFAR-FS} \cite{CIFAR}, with detailed results in Tables \ref{tab:results_fsc_MiniImagenet} and \ref{tab:results_fsc_Cifar} respectively. 
%For \emph{MiniImagenet} (Table \ref{tab:results_fsc_MiniImagenet}) we report on both versions \BPAp and \BPAt over a range of backbone architectures, while for the smaller datasets \emph{CIFAR-FS}  and \emph{CUB} (Table \ref{tab:results_fsc_Cifar}) we focus on the `drop-in' version \BPAp and only the strongest WRN architecture.
We evaluate the performance of the proposed BPA, applying it to a variety of FSC methods including the recent state-of-the-art (PTMap \cite{hu2020leveraging}, SillNet \cite{zhang2021sill}, PTMap-SF \cite{chen2021few} and PMF \cite{hu2022pushing}) as well as to conventional methods like the popular ProtoNet \cite{snell2017prototypical}. While in the \emph{MiniImagenet} evaluation we include a wide range of methods and backbones, in the \emph{CIFAR-FS} evaluation we focus on the state-of-the-art methods and configurations. %The detailed results are presented in Tables \ref{tab:results_fsc_MiniImagenet} and \ref{tab:results_fsc_CIFAR_CUB}) for the different datasets. 

For each evaluated 'hosting' method, we incorporate BPA into the pipeline as follows. Given an FSC instance, we transform the entire set of method-specific feature representations using BPA, in order to better capture relative information. The rest of the pipeline is resumed, allowing for both inference and training. Note that BPA flexibly fits into the FSC task, with no required knowledge or assumptions regarding the setting (\# of ways, shots or queries).

% \textbf{Notations.} All the results describe under the following section 'BPA insertion without network retraining'  notate as -\BPAp. The result described under 'BPA insertion with network retraining' notate as -\BPAt.

%, on the standard few-shot benchmarks MiniImagenet \cite{vinyals2016matching}, CIFAR-FS \cite{CIFAR}, and CUB \cite{CUB}
%, using different backbones. To do so, we transformed the original features with our BPA \simon{Daniel - what is "in the inception part"?} and reported the results on different configurations.

%=============================================================
The basic `drop-in' {\BPAp} consistently, and in many cases also significantly, improves the hosting method performance, including state-of-the-art, across all benchmarks and backbones with accuracy improvement of around $3.5\%$ and $1.5\%$ on the $1$- and $5$- shot tasks. 
This improvement without re-training the embedding backbone  shows BPA's effectiveness in capturing meaningful relationships between features in a very general sense. 
When re-training the hosting network with BPA inside, in an end-to-end fashion, {\BPAt} provides further improvements, in almost every method, with averages of  $5\%$ and $3\%$ on the $1$- and $5$- shot tasks. 

While most of the leading methods are transductive, our inductive version, \BPAi, can be seen to steadily improve on inductive methods like ProtoNet and PMF, without introducing transductive inference. This further emphasizes the generality and applicability of our method.

% Lastly, Table \ref{tab:results_fsc_MiniImagenet} additionally includes several interesting baselines to be pointed out. 
% The networks `PTMap-cosine' and `PTMap-softmax' (reported unsuccessful here) stand for two obvious baseline attempts (in this case building on PTMap~\cite{hu2020leveraging}) that work in the line of our approach, without the specialized OT-based transform. In the former, we take the output features to be the rows of the (un-normalized) matrix $\mathcal{S}$ (rather than those of $\mathcal{W}$) and in the latter we also normalize its rows using soft-max. 
%
% Second, the results the state-of-the-art self- (or co-) attention based methods ReNet~\cite{ReNet}, DeepEMD~\cite{DeepEMD} and FEAT~\cite{ye2020few}, being of close relation to ours, \simon{TBD}
%

%-----------------------------%-----------------------------
\begin{table}[t] \vspace{-12pt}
\centering
    \caption{
        \textbf{Unsupervised Image Clustering (UIC)} results on \emph{STL-10} \cite{coates2011analysis}, \emph{CIFAR-100-20} \cite{krizhevsky2009learning} and \emph{CIFAR-100-20} \cite{krizhevsky2009learning}.}
        \vspace{3pt}
    \fontsize{8.1}{8.1} \selectfont
    \setlength{\tabcolsep}{0.28em} % for the horizontal padding
    \renewcommand{\arraystretch}{1.2}
    \begin{tabular}{l | ccc | ccc | ccc}%{|l|l|c|}
    \hline
    \textbf{\textit{benchmark}} & \multicolumn{3}{c}{\emph{STL-10} %\cite{coates2011analysis}
    }  & \multicolumn{3}{|c}{\emph{CIFAR-10} %\cite{krizhevsky2009learning}
    }  & \multicolumn{3}{|c}{\emph{CIFAR-100-20} %\cite{krizhevsky2009learning}
    }  \\
    \hline    
    \textbf{\textit{network}}     &  
    ACC & NMI & ARI & ACC & NMI & ARI & ACC & NMI & ARI  \\
    \hline 
    k-means 
    & \scriptsize{0.192}&\scriptsize{0.125}&\scriptsize{0.061}&\scriptsize{0.229}&\scriptsize{0.087}&\scriptsize{0.049}&\scriptsize{0.130}&\scriptsize{0.084}&\scriptsize{0.028}\\
    DAC 
   &\scriptsize{0.470}&\scriptsize{0.366}&\scriptsize{0.257}&\scriptsize{0.522}&\scriptsize{0.396}&\scriptsize{0.306}&\scriptsize{0.238}&\scriptsize{0.185}&\scriptsize{0.088}\\
    DSEC 
   &\scriptsize{0.482}&\scriptsize{0.403}&\scriptsize{0.286}&\scriptsize{0.478}&\scriptsize{0.438}&\scriptsize{0.340}&\scriptsize{0.255}&\scriptsize{0.212}&\scriptsize{0.110}\\
    IDFD 
   &\scriptsize{0.756}&\scriptsize{0.643}&\scriptsize{0.575}&\scriptsize{0.815}&\scriptsize{0.711}&\scriptsize{0.663}&\scriptsize{0.425}&\scriptsize{0.426}&\scriptsize{0.264}\\ 
    SPICE$_s$
   &\scriptsize{0.908}&\scriptsize{0.817}&\scriptsize{0.812}&\scriptsize{0.838}&\scriptsize{0.734}&\scriptsize{0.705}&\scriptsize{0.468}&\scriptsize{0.448}&\scriptsize{0.294}\\
    SPICE 
   &\scriptsize{\it{\textbf{0.938}}}&\scriptsize{\it{\textbf{0.872}}}&\scriptsize{\it{\textbf{0.870}}}&\scriptsize{\it{\textbf{0.926}}}&\scriptsize{\it{\textbf{0.865}}}&\scriptsize{\it{\textbf{0.852}}}&\scriptsize{\it{\textbf{0.538}}}&\scriptsize{0.567}&\scriptsize{\it{\textbf{0.387}} 
   }\\
    \hdashline
    SPICE$_s$-\BPAts 
   &\scriptsize{0.912}&\scriptsize{0.823}&\scriptsize{0.821}&\scriptsize{0.880}&\scriptsize{0.784}&\scriptsize{0.769}&\scriptsize{0.494}&\scriptsize{0.477}&\scriptsize{0.334}\\
    SPICE-\BPAts 
   &\scriptsize{\textbf{0.943}}&\scriptsize{\textbf{0.880}}&\scriptsize{\textbf{0.879}}&\scriptsize{\textbf{0.933}}&\scriptsize{\textbf{0.870}}&\scriptsize{\textbf{0.866}}&\scriptsize{\textbf{0.550}}&\scriptsize{\it{\textbf{0.560}}}&\scriptsize{\textbf{0.402}}\\
    \hline           
    \end{tabular} \vspace{-11pt}
    \label{tab:image_clustering} 
\end{table}
%-----------------------------%-----------------------------

%=============================================================
\subsection{Unsupervised Image Clustering (UIC)} \label{sec.res.clustering}

Next, we evaluate BPA in the unsupervised domain, using the unsupervised image clustering task, with the additional challenge of capturing the relation between features that were learned without labels. 
To do so, we adopt SPICE \cite{niu2021spice}, a recent method that has shown phenomenal success in the field.
In SPICE, training is divided into 3 phases: (i) unsupervised representation learning (using MoCo \cite{he2019moco} over a resnet-34 backbone); (ii) clustering-head training, with result termed SPICE$_{s}$; and (iii) a joint training phase (using FixMatch \cite{sohn2020fixmatch} over a wrn backbone),  result termed SPICE. 

We insert BPA into phase (ii), clustering-head training, as follows. Given a batch of representations, SPICE assigns class pseudo-labels to the nearest neighbors of the most probable samples ($k$ samples with the highest probability per class). In the original work, SPICE uses the dot-product of the MoCo features to find the neighbors. Instead, we transform each batch of MoCo features using BPA and use the same dot-product on the resulting informative BPA features to find a more reliable set of neighbors.
We experiment on 3 standard datasets, \emph{STL-10} \cite{coates2011analysis}, \emph{CIFAR-10} and \emph{CIFAR-100-20} \cite{krizhevsky2009learning}, while keeping all original SPICE implementation hyper-parameters unchanged. We report both SPICE$_{s}$ and SPICE results, as in the original work \cite{niu2021spice}.

Table \ref{tab:image_clustering} summarizes the experiment, in terms of clustering Accuracy (ACC), Normalized Mutual Information (NMI), and Adjusted Rand Index (ARI). It is done for the two stages of SPICE, with and without BPA, along with several other baselines. The results show a significant improvement of SPICE$_s$-\text{{\BPAt}} over SPICE$_s$ (just by applying BPA to the learned features), with an average increase of $5\%$ in NMI and $8\%$ in ARI. The advantage brought by the insertion of BPA carries on to the joint-processing stage (\text{{\BPAt}} over SPICE$_s$), though with a smaller average increase of $0.1\%$ in NMI and $2.2\%$ in ARI, leading to new state-of-the-art results on these datasets.
These results demonstrate the relevance of BPA to unsupervised feature learning setups and its possible potential to other applications in this area.
\subsection{Person Re-Identification (Re-ID)}\label{sec.res.reid}
We explore the application of BPA to large-scale instances and datasets by considering the person re-identification task \cite{ye2021deep}. 
Given a set of \textit{query} images and a large set of \textit{gallery} images, the task is to rank the similarities of each query against the entire gallery. This is typically done by learning specialized image features that are compared by Euclidean distances. 
BPA is used to replace such pre-computed image features, by a well balanced representation with strong relative information, that is jointly computed over the union of query and gallery features. BPA is applied on pre-trained TopDBNet \cite{Top-DB-Net} resnet-50 features and tested on the large-scale ReID benchmarks \emph{CUHK03} \cite{CUHK03} (both 'detected' and `labeled') as well as the \textit{Market-1501} \cite{Market} set, reporting mAP (mean Average Precision) and Rank-1 metrics.

%-----------------------------%-----------------------------
\begin{table}[t] \vspace{-12pt}
\centering
    \caption{
        \textbf{Image Re-Identification (Re-ID)} results on \emph{CUHK03} \cite{CUHK03} and \emph{Market-1501} \cite{Market}.}  % \\  
    \vspace{3pt}
    \fontsize{8.1}{8.1} \selectfont
    \renewcommand{\arraystretch}{1.2}
    \setlength{\tabcolsep}{0.5em} % for the horizontal padding
    \begin{tabular}{l | cc | cc| cc}%{|l|l|c|}
    \hline
    \textit{\textbf{benchmark}} & \multicolumn{2}{c}{\emph{CUHK03-det}}
    & \multicolumn{2}{|c}{\emph{CUHK03-lab}
    % \cite{CUHK03}
    } & \multicolumn{2}{|c}{\emph{Market-1501} 
    % \cite{Market}
    }  \\
    \hline 
    \textit{\textbf{network}}    &  {~mAP}  &  {Rank-1} &  {mAP}  &  {Rank-1} &  {mAP}  &  {Rank-1}\\\hline
    MHN  & {~65.4}   &  {71.7}   &  {72.4}  &  {77.2}   & {85.0}  &  {95.1}\\
    SONA   & {~76.3}   &  {79.1}  &  {79.2}  &  {81.8}    & {88.6}   &  {95.6}\\
    OSNet  & {~67.8}   &  {72.3}  &  {--}  &  {--}  & {84.9}   &  {94.8}  \\
    Pyramid  & {74.8}   &  {78.9}   &  {76.9}  &  {81.8}   & {88.2}   &  {\textbf{95.7}}\\
    TDB  & {~72.9}   &  {75.7}   &  {75.6}  &  {77.7}    & {85.7}   &  {94.3}\\
    TDB$_{RK}$ & {\it{\textbf{87.1}}}   &     {\it{\textbf{87.1}}}     &  \it{\textbf{89.1}}  &  \it{\textbf{89.0}}    &   {\textbf{94.0}}   &    {\it{\textbf{95.3}}}   \\
    \hdashline
    TDB-\BPAps  & {~{{77.9}}}  &  {80.4} &  {80.4}  &  {82.6}   &     {\textbf{\it{88.1}}}  &   {{94.4}} \\
    TDB$_{RK}$-\BPAps     &   {~\textbf{87.9}}    &         {\textbf{{88.0}}} &  {\textbf{89.5}}  &  {\textbf{89.8}}  &  {\textbf{94.0}}  &  {{95.0}}\\
    \hline           
    \end{tabular} \vspace{-9pt}
    \label{tab:results_on_DukeMTMC_and_market}
\end{table}
%-----------------------------%-----------------------------

In Table \ref{tab:results_on_DukeMTMC_and_market}, TDB and TDB$_{RK}$ are shorthands for using  TopDBNet features, before and after re-ranking \cite{Re-ranking}. There is a consistent benefit in applying BPA to these state-of-the-art features, prior to the distance computations, with a significant average increase of over $5\%$ in mAP and $4\%$ in Rank-1 prior to re-ranking and a modest increase of $0.5\%$ in both measures after ranking. %, leading to new state-of-the-art results.
These results demonstrate that BPA can handle large-scale instances (with thousands of features) and successfully improve performance measures in such retrieval oriented tasks. %\simon{leave this?: ReID was not the main focus of this work, hence, we did not re-train the hosting networks with BPA included. Further experimentation is required to measure the possible effects of doing so.}

% <<<<<<<<<<<<<<<<<<<<<<<<<

%-------------------------------------------------------------------
% conclusions
%-------------------------------------------------------------------
\section{Conclusions, Limitations and Future Work}

We presented a novel feature-embedding approach for set-input grouping-related tasks such as clustering, classification and retrieval. The proposed BPA feature-set transform is non-parametric, differentiable, efficient, easy to use and is shown to capture complex relations between the set-input items.
Applying BPA to the tasks of few-shot-classification, unsupervised-image-clustering and person-re-identification, whether by insertion into a pre-trained network or by re-training the hosting network, has shown across-the-board improvements, setting new state-of-the-art results. 
%Based on these promising results, \hl{we believe that} understanding the full potential of the transform and its resulting representation requires further studying.

In future work, we plan to address current limitations and explore potential extensions. 
%Regarding the output dimensionality of the embedding, which is dictated by the input set size, we aim to obtain arbitrary dimensions, for further increased usage flexibility. Initial results using PCA (without supporting analysis) are showing promising results. 
BPA is currently limited to producing features that represent \textit{relative} information, within the set-items. It could possibly be applied to tokens (e.g. patches) of a single item (e.g. image), similar to transformers, perhaps dropping the equivariance property and utilizing spatial encoding, to improve non-relative representations. In addition, it could be useful for guiding contrastive self-supervised learning, where embeddings are trained by relative information of augmented views.   % <<<<<<<<<<<<<<<<<<<<<<<<<

% %-------------------------------------------------------------------
% % Impact Statement
% %-------------------------------------------------------------------
% \section*{Impact Statement}

% This paper presents work whose goal is to advance the field of 
% Machine Learning. There are many potential societal consequences 
% of our work, none which we feel must be specifically highlighted here.

%-------------------------------------------------------------------
% refs
%-------------------------------------------------------------------
%\newpage

\bibliography{reference}
\bibliographystyle{icml2024}

%%%%%%%%%%%%%%%%%%%%%%%%%%%%%%%%%%%%%%%%%%%%%%%%%%%%%%%%%%%%%%%%%%%%%%%%%%%%%%%
%%%%%%%%%%%%%%%%%%%%%%%%%%%%%%%%%%%%%%%%%%%%%%%%%%%%%%%%%%%%%%%%%%%%%%%%%%%%%%%
% APPENDIX
%%%%%%%%%%%%%%%%%%%%%%%%%%%%%%%%%%%%%%%%%%%%%%%%%%%%%%%%%%%%%%%%%%%%%%%%%%%%%%%
%%%%%%%%%%%%%%%%%%%%%%%%%%%%%%%%%%%%%%%%%%%%%%%%%%%%%%%%%%%%%%%%%%%%%%%%%%%%%%%

% \clearpage
\appendix
% \onecolumn

\vspace{12pt}
{\fontsize{18.1}{18.1} \selectfont {\textbf{Appendix}}}
\vspace{4pt}

{\fontsize{11}{11} \selectfont
The Appendix includes the following sections:
\vspace{2pt}

\hspace{8pt}\hyperref[sec:BPA_impl]{A. PyTorch-style BPA Implementation} 

\hspace{8pt}\hyperref[sec:ablations]{B. Ablation Studies} 

\hspace{8pt}\hyperref[sec:insertions]{C. BPA Insertion into Hosting Algorithms} 

\hspace{8pt}\hyperref[sec:case_study]{D. {Clustering on the Sphere - a Case Study} 
}}\vspace{2pt}

%-------------------------------------------------------------------
% BPA pseudocode
%-------------------------------------------------------------------
\section{PyTorch-style BPA Implementation} \label{sec:BPA_impl}
%~~~~~~~~~~~~~~~~~
%In the main paper, the proposed BPA feature transform was fully derived (Sec. 3.1), analyzed (Secs. 3.1, 3.2) and illustrated (Fig. 2). 
We provide in Algorithm \ref{algo:BPA} a PyTorch Style implementation that fully aligns with the description in the paper as well as with our actual implementation that was used to execute all of the experiments. In Appendix \ref{sec:insertions} we further demonstrate the "insertions" of BPA into hosting methods, for each of our three main applications.

\underline{Notice mainly that}: 
(i) The transform can easily be dropped-in, using the simple one-line call: {X =  {\color{purple}{\textbf{BPA}}}(X)}. 
(ii) It is fully differentiable (as Sinkhorn and the other basic operations are). 
(iii) The transform does not need to know (or even assume) anything about the number of features, their dimension, or distribution statistics among classes (e.g. whether balanced or not).

It follows the \underline{simple steps of}: (i) Computing Euclidean self pairwise distances (using cosine similarities between unit normalized input features); (ii) Avoiding self-matching by placing infinity values on the distances matrix diagonal; (iii) Applying a standard Sinkhorn procedure, given the distance matrix and the only 2 (hyper-) parameters with their fixed values: entropy regularization parameter $\lambda$ and the number of row/col iterative normalization steps. Note that Sinkhorn defaultly maps between source and target vectors of ones; (iv) Restoring the perfect self-matching probabilities of one, along the diagonal.
 
%%%%%%%%%%%%%%%%%%%%%%%%%%%%%%%%%%%%%%%%%%%%%%%%%%%%%%%%%%%%%%%%%%
\definecolor{commentcolor}{RGB}{0,140,0}   % define comment color
\newcommand{\PyComment}[1]{{\small\ttfamily\textcolor{commentcolor}{\# #1}}}  % add a "#" before the input text "#1"
\newcommand{\PyCode}[1]{{\small\ttfamily\textcolor{black}{#1}}} % \ttfamily is the code font
  \vspace{10pt}
% \LinesNumbered
\setlength{\textfloatsep}{0.25cm}
% \setlength{\floatsep}{0.1cm}
%%%%%%%%%%%%%%%%%%%%%%%%%%%%%%%%%%%%%%%%%%%%%%%%%%%%%%%%%%%% 1 %%%
% \begin{figure}
% \vspace{55pt}  
\begin{algorithm}[b!]
\normalsize
% \SetAlgoLined
% \setlength{\algomargin}{5.5em}
    % \PyCode{\textbf{def} {\color{purple}{BPA}}(V):}\\
     %
  % \Indp   % start indent
  \vspace{3pt}
  \hspace{-1pt}
    % \PyComment{compute self pairwise-distances}\\\vspace{1pt}\hspace{4pt}
    \PyCode{\textbf{def} {\color{purple}{\textbf{BPA}}}(V):}\\ \vspace{2pt}
  \hspace{4pt}
    \PyComment{compute self pairwise-distances}\\\vspace{1pt}\hspace{7pt}
    \PyCode{D = 1 - pwise\_cosine\_sim(V/V.norm())}\\ \vspace{2pt}
  \hspace{4pt}
    \PyComment{infinity self-distances on diagonal}\\\vspace{1pt}\hspace{7pt}   
    \PyCode{D\_inf = D.fill\_diagonal(10e9)}\\ \vspace{2pt}
  \hspace{4pt}
    \PyComment{compute optimal transport plan}\\\vspace{1pt}\hspace{7pt}
    \PyCode{W = Sinkhorn(D\_inf,lambda=.1,iters=5)}\\ \vspace{2pt}
  \hspace{4pt}
    \PyComment{stretch affinities to [0,1]}\\\vspace{1pt}\hspace{7pt}
    \PyCode{W = W/W.max()}\\ \vspace{2pt}
     %
  % \hspace{1pt}
  %   \PyComment{divide the BPA matrix by its max}\\\hspace{4pt}
  %   \PyCode{W = W / W.flatten().max()}
     %
  \hspace{4pt}
    \PyComment{self-affinity on diagonal to 1}\\\vspace{1pt}\hspace{7pt}
    \PyCode{return W.fill\_diagonal(1)}

  \vspace{-7pt}\hspace{7pt}
  % \Indm % end indent

\caption{\textbf{{\color{purple}BPA}} transform on a set of $n$ features. \\ 
\textbf{input}: $n\times d$ matrix $V$ $\quad$ \textbf{output}: $n\times n$ matrix $W$}
\label{algo:BPA} 
\end{algorithm}
% \end{figure}
% \setlength{\textfloatsep}{0.8cm}
%%%%%%%%%%%%%%%%%%%%%%%%%%%%%%%%%%%%%%%%%%%%%%%%%%%%%%%%%%%%%

 %%%%

%-------------------------------------------------------------------
% ablations
%-------------------------------------------------------------------
\section{Ablation Studies} \label{sec:ablations}

%-----------------------------------------------------------------
\subsection{Scalability (accuracy, runtime vs. input size)}
%-----------------------------------------------------------------
Being a transductive module, the accuracy and efficiency of the BPA transform depend on the number of inputs that are processed as a batch. Recall that BPA is a drop-in addition that usually follows feature extraction and precedes further computation - e.g. $k$-means for clustering, or (often transductive) layers in FSC and ReID. 

The ReID experiment is a good stress-test for BPA, since we achieve excellent results for batch sizes of up to $\sim\hspace{-4pt}$ 15K image descriptors.
In terms of runtime, although BPA's complexity is quadratic in sample size, its own (self) runtime is empirically negligible compared to that of the processing that follows, in all applications tested. %In fact, its only affect on the runtime is through increasing the feature dimensionality (for very large batches), slightly affecting the further processing stages.

\begin{figure}[t!] \vspace{-3pt}
\centering
 \includegraphics[width=1.02\columnwidth]{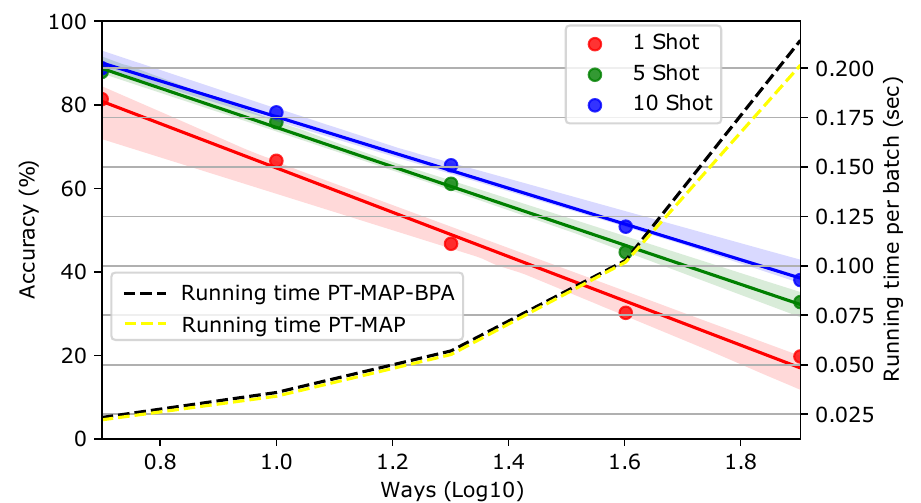} \\
     \vspace{-7pt}\caption{
   \textbf{BPA scaling in terms of accuracy and efficiency}. %See text for details.
   }
 \label{fig.scaling}\vspace{4pt}
\end{figure}

Typical FSC tasks sizes ($(shots+queries)\cdot ways$) are small: $100=(5+15)\cdot 5$ at the largest. To concretely address this matter, we test a resnet-12 PTMap-{\BPAp} on large-scale FSC, following \cite{dhillon2019baseline}, on the Tiered-Imagenet dataset and report accuracy for 1/5/10-shot (15-query) tasks for an increasing range of ways. The results, shown in Fig.~\ref{fig.scaling}, show that: (i) Total runtime, where BPA is only a small contributor (compare black vs. yellow dashed line), increases gracefully (notice log10 x-axis) even for extremely large FSC tasks of $4000=(10+15)\cdot 160$ images; (ii) Our accuracy scales as expected - following the observation in \cite{dhillon2019baseline} that it changes logarithmically with ways (straight line in log-scale).

%% --%%%%%%%%%%%%%%% figure SYNTHETIC %%%%%%%%%%%%%%%%%%----
%% --%%%%%%%%%%%%%%%%%%%%%%%%%%%%%%%%%%%%%%%%%%%%%%%%%%%----

%% -------------------- figure LAMBDA FSC ---------------
\begin{figure}[t]
\centering    
\vspace{-3pt}
\setlength{\tabcolsep}{0.53em} % for the horizontal padding
     \includegraphics[width=\columnwidth]{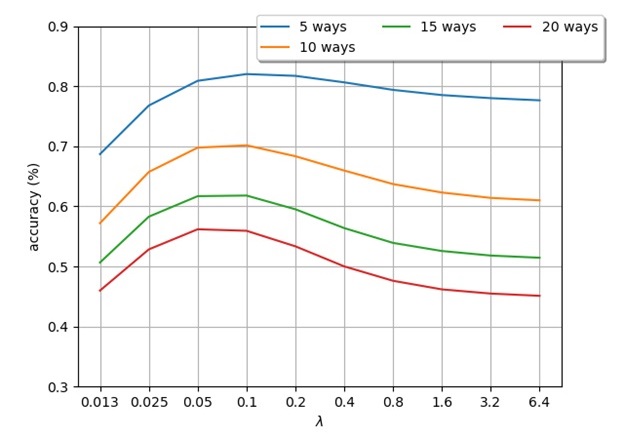}\\ \vspace{-5pt}    
     \includegraphics[width=\columnwidth]{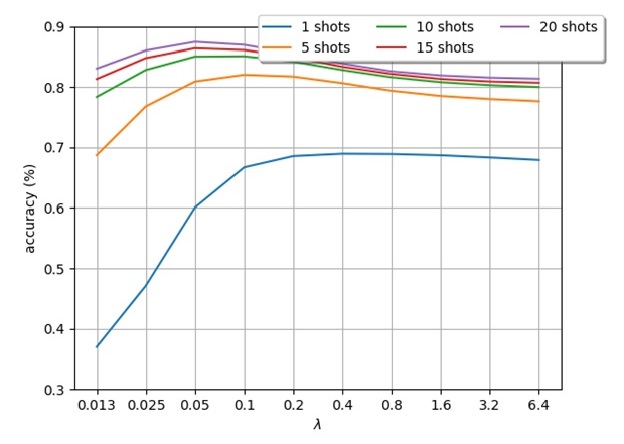}\\
  \vspace{-14pt}\caption{
    \textbf{Ablation of entropy regularization parameter $\lambda$ using the \textit{Few-Shot-Classification} (FSC) task}: Considering different `ways' (\textbf{top}), and different `shots' (\textbf{bottom}). See text for details.
    }
     \label{fig.lambda_exp}\vspace{8pt}
\end{figure}
%% ---------------------------------------------------

\subsection{Sinkhorn Iterations}
In Table \ref{tab:results_sinkhorn_iters} we ablate the number of normalization iterations in the Sinkhorn-Knopp (SK) \cite{cuturi2013sinkhorn} algorithm at test-time. We measured accuracy on the validation set of \emph{MiniImagenet} \cite{vinyals2016matching}, using ProtoNet-{\BPAp} (which is the non-fine-tuned drop-in version of BPA within ProtoNet \cite{snell2017prototypical}). 
As was reported in prior works following \cite{cuturi2013sinkhorn}, we empirically observe that a very small number of iterations provide rapid convergence, with diminishing return for higher numbers of iterations. We observed similar behavior for other hosting methods, and therefore chose to use a fixed number of 5 iterations throughout the experiments.

%----------------------------- TABLE FSC: SinkhornIterations -------------
\begin{table}[h!]
\normalsize
 \caption{
      {
\textbf{Sinkhorn iterations ablation study:} See text for details.}}
    \centering\vspace{6pt}
    \setlength{\tabcolsep}{0.39em} % for the horizontal padding
    \begin{tabular}{cccc}%{|l|l|c|c|}
    \hline    
    {\textbf{method}}    &  
    {\textbf{iters}}  &  
    \wayshot{5}{1}  &  
    \wayshot{5}{5}   \\  \hline 
    ProtoNet-{\BPAp}      &              1              &            70.71            &       83.79 \\ 
    ProtoNet-{\BPAp}      &              2              &            71.10            &       84.01 \\ 
    ProtoNet-{\BPAp}      &              4              &            71.18            &       84.08 \\ 
    ProtoNet-{\BPAp}      &              8              &            71.20            &       84.10 \\ 
    ProtoNet-{\BPAp}      &              16             &            71.20            &       84.10 \\\hline   
    \end{tabular} \vspace{5pt}
    \vspace{-5pt}
    \label{tab:results_sinkhorn_iters}
\end{table}
%-------------------------------------------------------------

% \centering\vspace{-2pt}
   %  \footnotesize
   %  \setlength{\tabcolsep}{0.43em} % for the horizontal padding
   %  \renewcommand{\arraystretch}{1.2}
   %  \begin{tabular}{lcccc}%{|l|l|c|c|}
   %  \hline    
   %  {\textbf{method}}    &  {\textbf{\textit{T/I}}}    &  {\textbf{backbone}}  &  \wayshot{5}{1}  &  \wayshot{5}{5}   \\\hline 

\subsection{Sinkhorn Entropy Regularization $\lambda$} \label{sec.lambda}
\begin{figure}[t!]
\vspace{-6pt}
\includegraphics[width=\columnwidth]{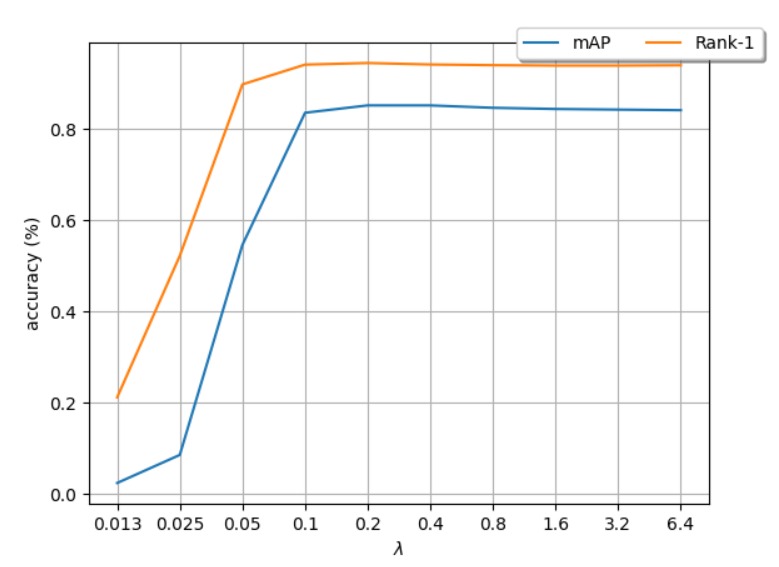}
        \\ \vspace{-36pt} \\
    \caption{
   \textbf{Ablation of entropy regularization parameter $\lambda$ using the \textit{Person-Re-Identification} (Re-ID) task}. Accuracy vs. $\lambda$, using the validation set of Market-1501 \cite{Market} and considering both mAP and Rank-1 measures. See text for details.
   }
\label{fig.lambda_exp_reid}
\vspace{12pt}
\end{figure}

We measured the impact of using different values of the optimal-transport entropy regularization parameter $\lambda$ (the main parameter of the Sinkhorn algorithm) on a variety of configurations (ways and shots) in Few-Shot-Classification (FSC) on \emph{MiniImagenet} \cite{vinyals2016matching} in Fig. \ref{fig.lambda_exp} as well as on the Person-Re-Identification (RE-ID) experiment on Market-1501 \cite{Market} in Fig. \ref{fig.lambda_exp_reid}. In both cases, the ablation was executed on the validation set.

For FSC, in Fig. \ref{fig.lambda_exp}, the \textbf{top} plot shows that the effect of the choice of $\lambda$ is similar across tasks with a varying number of ways. The \textbf{bottom} plot shows the behavior as a function of $\lambda$ across multiple shot-values, where the optimal value of $\lambda$ can be seen to have a certain dependence on the number of shots. Recall that we chose to use a fixed value of $\lambda=0.1$, which gives an overall good accuracy trade-off. Note that a further improvement could be achieved by picking the best values for the particular cases. Notice also the log-scale of the x-axes to see that performance is rather stable around the chosen value.

% %% -------------------- figure LAMBDA FSC ---------------
% \begin{figure}[h!]
% \centering    
% \includegraphics[width=\columnwidth]{figures/supplementary/lambda_exp_part1.jpg}\\ \includegraphics[width=\columnwidth]{figures/supplementary/lambda_exp_part2.jpg}\\
%   \\\vspace{-6pt}\caption{
%     \textbf{Ablation study on $\lambda$ in \textit{Few-Shot-Classification} (FSC)}: Considering different `ways' (left), and different `shots' (right). See text for details.
%     }
%      \label{fig.lambda_exp}\vspace{-3pt}
% \end{figure}
% %% ---------------------------------------------------

%
%
%
%% ------------------------- figure lambda ReID  -----------

For Re-ID, in Fig. \ref{fig.lambda_exp_reid}, we experiment with a range of $\lambda$ values on the validation set of the Market-1501 dataset. The results (shown both for mAP and rank-1 measures) reveal a strong resemblance to those of the FSC experiment in Fig. \ref{fig.lambda_exp}, however, the optimal choices for $\lambda$ are slightly higher, which is consistent with the dependence on the shots number, since the re-ID tasks are typically large ones. We found that a value of $\lambda=0.25$ gives good results across both datasets.

\subsection{BPA vs. Naive Baselines} \label{sec.baseline}

\begin{figure}[b!]
\centering    \vspace{8pt}
     % \includegraphics[width=\columnwidth]{figures/supplementary/baselines_vs_ways.png}
     % \\
     \includegraphics[width=0.9\columnwidth]{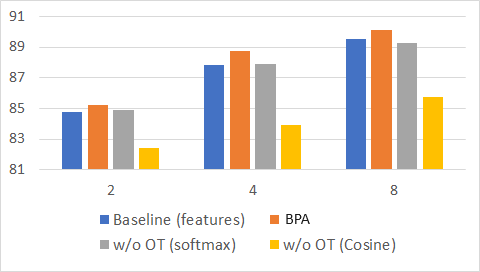}\\
   %   \vspace{-13pt}\\
   % (a) accuracy vs. baselines over different 'ways' (top) and over different 'shots' (bottom)\\
  \vspace{-5pt}\caption{
    \textbf{Comparison of BPA to different baselines over different configurations} in few-shot learning tasks over \emph{MiniImagenet} \cite{vinyals2016matching}. Created by measuring accuracy ($y$-axis) over a varying number of shots ($x$-axis), with fixed 5-ways and 15-queries. See text for details.
    }
     \label{fig.baselines_exp}\vspace{-10pt}
\end{figure}

% Within the context of few-shot learning on \emph{MiniImagenet}, we performed several ablation studies. 
In Fig. \ref{fig.baselines_exp}, we ablate different simple alternatives to BPA, with the PTMap \cite{hu2020leveraging} few-shot-classifier as the 'hosting' method, using \emph{MiniImagenet} \cite{vinyals2016matching}. Each result is the average of 100 few-shot episodes, using a WRN-28-10 backbone feature encoder. 
In blue is the baseline of applying no transform at all, using the original features. In orange - using BPA. In gray and yellow, respectively, are other naive ways of transforming the features, where the affinity matrix is only row-normalized ('softmax') or not normalized at all ('cosine') before taking its rows as the output features. It is empirically evident that only BPA outperforms the baseline consistently, which is due to the properties that we had proved regarding the transform.
%% ---------------------------------------------------

\vspace{3pt}
\section{BPA Insertion into Hosting Algorithms} \label{sec:insertions}
%~~~~~~~~~~~~~~~~~

\vspace{3pt}
\subsection{PTMap \hspace{-1pt}\cite{hu2020leveraging} (\textbf{Few-Shot Classification})}
\vspace{-2pt}
%~~~~~~~~~~~~~~~~~
We present the pseudo-code for utilizing BPA within the PTMap pipeline, as outlined in Alg. \ref{algo_PTMAP}. The only alteration from the original implementation pertains to row 5, wherein the support and query sets are concatenated and transformed using BPA. This approach can be extended to a wide range of distance-based methodologies, thus providing a simple and versatile solution to a variety of applications.

%%%%%%%%%%%%%%%%%%%%%%%%%%%%%%%%%%%%%
%%%%%%%%%%%%%%%%%%%%%%%%%%%%%%%%%%%%%%%%%%%%%%%%%%%%%%%%%%%%%%%%%%
\definecolor{commentcolor}{RGB}{0,140,0}   % define comment color
\renewcommand{\PyComment}[1]{{\footnotesize\ttfamily\textcolor{commentcolor}{\# #1}}}  % add a "#" before the input text "#1"
\renewcommand{\PyCode}[1]{{\footnotesize\ttfamily\textcolor{black}{#1}}} % \ttfamily is the code font

\setlength{\textfloatsep}{0.25cm}
% \setlength{\floatsep}{0.1cm}
%%%%%%%%%%%%%%%%%%%%%%%%%%%%%%%%%%%%%%%%%%%%%%%%%%%%%%%%%%%% 1 %%%
% \vspace{-7pt}
% \begin{figure}

% ======================== PT-MAP =================
\begin{algorithm}[h!]
% \begin{algorithmic}[1]
\small
\label{algo_PTMAP}
%\SetAlgoLined
\vspace{2pt}
\caption{ \textbf{{\color{teal}PTMap} \textit{training} and \textit{inference}}}
\hspace{-1pt}\vspace{3pt}
\textbf{inputs:} $\mathbf{x_s},\mathbf{x_q}$ \hspace{2pt}
\PyComment{support, query images}\\ \vspace{-1pt}
\hspace{27pt} $\ell_\mathbf{s}, (\ell_\mathbf{q}$) \hspace{-2pt}\vspace{3pt}
\PyComment{support, (query) labels}\\ \vspace{-2pt} 
\hspace{27pt} $f_{\phi}$ \hspace{16pt}
\PyComment{pre-trained embedding network}\\ \vspace{0pt}

%--------
\hspace{-2pt}\vspace{3pt}
$\mathbf{f_s}=f_{\phi}(\mathbf{x_s}), \:\mathbf{f_q}=f_{\phi}(\mathbf{x_q})$  \hspace{0pt}
\PyComment{extract features}\\ \vspace{2pt}
\hspace{-3pt}
$(\mathbf{f_s}\cup\mathbf{f_q})=\:$\PyCode{{\color{purple}{\textbf{BPA}}}}\hspace{1pt}$(\mathbf{f_s}\cup\mathbf{f_q})$ \hspace{0pt}
\PyComment{BPA transformed features}\\ \vspace{3pt}
\hspace{-3pt}
$\mathbf{c}_j= \frac{1}{s}\cdot\sum_{\mathbf{f}\in\mathbf{f_s},\ell_\mathbf{s}(\mathbf{f})=j}\mathbf{f}, \forall j$ \hspace{0pt}
\PyComment{init class centers}\\ \vspace{2pt}
\hspace{-3pt}
\textit{\textbf{repeat}:}\\\vspace{2pt}
    \hspace{0pt}
    $\boldsymbol{\cdot}\:$    $\mathbf{L}_{ij}=\|\mathbf{f}_i-\mathbf{c}_j\|^2, \;\forall i,\mathbf{f}_i\hspace{-2pt}\in\hspace{-2pt}\mathbf{f}_q$ \hspace{-1pt}
    \PyComment{\hspace{-2pt}feature-center dists}\\\vspace{3pt}
    \hspace{0pt}
    $\boldsymbol{\cdot}\:$        $\mathbf{M}=Sinkhorn(\mathbf{L}, \lambda)$ \hspace{-1pt}
    \PyComment{\hspace{-2pt}S-horn soft assignments}\\ \vspace{2pt}
    \hspace{0pt}
    $\boldsymbol{\cdot}\:$   $\mathbf{c}_j \leftarrow \mathbf{c}_j + \alpha (g(\mathbf{M},j) - \mathbf{c}_j), \;\forall j$ \hspace{-1pt}
    \PyComment{\hspace{-2pt}update centers}\\\vspace{4pt}
\hspace{-3pt}
$\hat{\ell_\mathbf{q}}(\mathbf{f}_i)=\arg\max_j(\mathbf{M}[i,j])$
\hspace{-3pt}
\PyComment{prediction per $\mathbf{f}_i\hspace{-2pt}\in\hspace{-2pt}\mathbf{f}_q$}\\\vspace{1pt}
%\vspace{-.5cm}
%
\textbf{if} \textit{inference}: \\\vspace{3pt}
    \hspace{11pt}\textbf{return} $\hat{\ell}_\mathbf{q}$ \hspace{0pt}
    \PyComment{\hspace{-1pt}query predictions}\\\vspace{2pt}
\textbf{else} (\textit{training}):\\\vspace{3pt}
    \hspace{8pt} \textbf{update} $f_{\phi}$ by $\nabla_{\phi}$C-Entropy($\mathbf{M}$, $\ell_\mathbf{q}$) \hspace{0pt}\PyComment{\hspace{-1pt}grad-desc.} \vspace{-6pt} \\
% \end{algorithmic}
\end{algorithm}
% ======================== PT-MAP ================= %%%%
%%%%%%%%%%%%%%%%%%%%%%%%%%%%%%%%%%%%%

\vspace{3pt}
\subsection{SPICE \cite{niu2021spice}  (\textbf{Unsupervised Clustering})}
\vspace{-2pt}
%~~~~~~~~~~~~~~~~~
In our implementation of SPICE, as detailed in the paper, we utilize BPA during phase 2 of the algorithm (clustering-head training). Specifically, as depicted in Alg. \ref{algo_SPICE_training}, we transform the features using BPA, batch-wise, before conducting a nearest-neighbor search. Afterwards, we retrieve the pseudo-labels and resume with the original features, as in the original implementation.

%%%%%%%%%%%%%%%%%%%%%%%%%%%%%%%%%%%%%
%%%%%%%%%%%%%%%%%%%%%%%%%%%%%%%%%%%%%%%%%%%%%%%%%%%%%%%%%%%%%%%%%%
\definecolor{commentcolor}{RGB}{0,140,0}   % define comment color

\setlength{\textfloatsep}{0.25cm}
% \setlength{\floatsep}{0.1cm}
%%%%%%%%%%%%%%%%%%%%%%%%%%%%%%%%%%%%%%%%%%%%%%%%%%%%%%%%%%%% 1 %%%
\vspace{-5pt}
% \begin{figure}

% % ======================== SPICE =================
% % \LinesNumberedHidden{

% \begin{algorithm}[h!]
% \small
% \label{algo_SPICE_clustering}
% \SetAlgoLined
% \SetKwInOut{Input}{inputs}
% \caption{SPICE [\textcolor{green}{28}] \textit{training} - Phase (ii)}
% %
% \hspace{-3pt}\vspace{3pt}
% \textbf{train clustering network $c_\theta$, given pre-trained embedding network $f_{\phi}$} by repeating per batch $\mathbf{x}$:
% \\
% %
% \hspace{1pt}\vspace{3pt}
% $\boldsymbol{\cdot}\:$ $\mathbf{f}=f_{\phi}(\mathbf{x})$ 
% $\quad$\PyComment{extract features}\\ \vspace{2pt}
% %
% \hspace{1pt}\vspace{3pt}
% $\boldsymbol{\cdot}\:$ $\mathbf{f}^\text{\color{purple}{BPA}}=\:$\PyCode{\color{purple}{BPA}}\hspace{1pt}$(\mathbf{f})$
% $\quad$\PyComment{BPA transformed features}\\
% %
% \hspace{1pt}\vspace{3pt}
% $\boldsymbol{\cdot}\:$ find 3 most confident samples per cluster (use $\mathbf{f}$)\\
% %
% \hspace{1pt}\vspace{3pt}
% $\boldsymbol{\cdot}\:$ compute cluster centers as their means (use $\mathbf{f}^\text{\color{purple}{BPA}}$)\hspace{-44pt}\\
% %
% \hspace{1pt}\vspace{3pt}
% $\boldsymbol{\cdot}\:$ find nearest-neighbors of each center (use $\mathbf{f}^\text{\color{purple}{BPA}}$)\hspace{-44pt}\\
% %
% \hspace{1pt}\vspace{3pt}
% $\boldsymbol{\cdot}\:$ assign them to the cluster (as pseudo-labels)\hspace{-94pt}\\
% %
% \hspace{1pt}\vspace{3pt}
% $\boldsymbol{\cdot}\:$ use pseudo-labels to train (update) the network $c_\theta$
% \\
% %\vspace{-.5cm}
% \end{algorithm}

% % }
% % ======================== SPICE =================

% ======================== SPICE =================
% \LinesNumberedHidden{
\begin{algorithm}[h!]
\label{algo_SPICE_training}
% \small
%\SetAlgoLined
\caption{\textbf{ {\color{teal}SPICE} \textit{training}}}
\vspace{4pt} 
{\textbf{Phase (i): pre-train embedding network $f_{\phi}$ }}
\vspace{3pt} \\
{\textbf{Phase (ii): train clustering network $c_\theta$}}
\hspace{6pt}\vspace{3pt}\\\hspace{6pt}
\textit{\textbf{repeat}} per batch $\mathbf{x}$:\\\vspace{2pt}
\hspace{0pt}
$\boldsymbol{\cdot}\:$ \hspace{-5pt} $\mathbf{f}=f_{\phi}(\mathbf{x})$ 
\PyComment{extract features}\\\vspace{2pt}
\hspace{0pt}
$\boldsymbol{\cdot}\:$ \hspace{-2pt}$\mathbf{f}^\text{\color{purple}{BPA}}=\:$\PyCode{\color{purple}{\textbf{BPA}}}\hspace{1pt}$(\mathbf{f})$
\PyComment{BPA transformed features}\\\vspace{2pt}
\hspace{0pt}
$\boldsymbol{\cdot}\:$ \hspace{-2pt}Find 3 most confident samples per cluster (use $\mathbf{f}$)\\\vspace{2pt}
\hspace{0pt}
$\boldsymbol{\cdot}\:$ \hspace{-2pt}Compute cluster centers as their means (use $\mathbf{f}^\text{\color{purple}{BPA}}$)\hspace{-44pt}\\\vspace{2pt}
\hspace{0pt}
$\boldsymbol{\cdot}\:$ \hspace{-2pt}Find nearest-neighbors of each center (use $\mathbf{f}^\text{\color{purple}{BPA}}$)\hspace{-44pt}\\\vspace{2pt}
\hspace{0pt}
$\boldsymbol{\cdot}\:$ \hspace{-2pt}Assign them to the cluster (as pseudo-labels)\hspace{-94pt}\\\vspace{2pt}
\hspace{0pt}
$\boldsymbol{\cdot}\:$ \hspace{-2pt}Use pseudo-labels to train (update) $c_\theta$
\vspace{4pt} \\
\hspace{-3pt}\vspace{3pt}
{\textbf{Phase (iii): jointly fine-tune $f_{\phi}$ and $c_\theta$}} \\
\vspace{-.3cm}
\end{algorithm}
% }
% ======================== SPICE ================= %%%%
%%%%%%%%%%%%%%%%%%%%%%%%%%%%%%%%%%%%%

\vspace{5pt}
\subsection{TopDBNet \cite{Top-DB-Net} (\textbf{Person ReID})}
\vspace{-2pt}
%~~~~~~~~~~~~~~~~~

Finally, Alg. \ref{algo:TDB} illustrates the application of BPA during inference in the context of Person ReID. Typically, the query identity search within the gallery involves identifying the nearest sample to each query. In our implementation, we adopt the same methodology, with the additional step of transforming the concatenated set of query and gallery features, using the BPA transform prior to the search. 

%%%%%%%%%%%%%%%%%%%%%%%%%%%%%%%%%%%%%
%%%%%%%%%%%%%%%%%%%%%%%%%%%%%%%%%%%%%%%%%%%%%%%%%%%%%%%%%%%%%%%%%%
\definecolor{commentcolor}{RGB}{0,140,0}   % define comment color

\setlength{\textfloatsep}{0.25cm}
% \setlength{\floatsep}{0.1cm}
%%%%%%%%%%%%%%%%%%%%%%%%%%%%%%%%%%%%%%%%%%%%%%%%%%%%%%%%%%%% 1 %%%
% \vspace{-7pt}
% \begin{figure}

% ======================== TDB =================
% \LinesNumberedHidden{
\begin{algorithm}[h!]
% \LinesNumbered
% \ShowLn
\small
\label{algo:TDB}
% \setalgolined
\caption{\hspace{0.01cm} \textbf{{\color{teal}TopDBNet} \textit{inference}}} 
\vspace{3pt}\hspace{-3pt}
\textbf{inputs:} $\mathbf{x_g},\mathbf{x_q}$ \hspace{1pt}
\PyComment{gallery images, query images}\\ \vspace{-2pt}
\hspace{24pt} $f_{\phi}$ \hspace{17pt}
\PyComment{pre-trained embedding network}\\ \vspace{0pt}

\PyComment{extract features}\vspace{3pt}\\ 
\hspace{-0pt}
\PyCode{$\mathbf{f_g}=f_{\phi}(\mathbf{x_g}), \:\mathbf{f_q}=f_{\phi}(\mathbf{x_q})$} \hspace{2pt}\vspace{-4pt}\\ 

\PyComment{transform them with BPA}\vspace{3pt}\\ 
\hspace{-0pt}
\PyCode{$(\mathbf{f_g}\cup\mathbf{f_q})=\:$\PyCode{{\color{purple}{\textbf{BPA}}}}\hspace{1pt}$(\mathbf{f_g}\cup\mathbf{f_q})$}  \hspace{2pt}\vspace{-4pt}\\ 

\PyComment{return gallery image with closest feature}\vspace{3pt}\\ 
\hspace{-0pt}
\PyCode{\textbf{return} $\underset{\{j:\mathbf{f}_j\in \mathbf{f_g}\}}{\mathrm{argmin}}$ $\|\mathbf{f}_i-\mathbf{f}_j\|$ for every $\{i:\mathbf{f}_i\in \mathbf{f_q}\}$} \hspace{2pt}\\
\end{algorithm}
% }
% ======================== TDB =================

% % ======================== TDB =================
% % \LinesNumberedHidden{
% \begin{algorithm}[h!]
% \small
% \label{algo:TDB}
% \SetAlgoLined
% \caption{ \textbf{TopDBNet \cite{Top-DB-Net} \textit{inference}}}
% %
% \hspace{-0pt}\vspace{3pt}
% \textbf{inputs:} $\mathbf{x_g},\mathbf{x_q}$ \hspace{1pt}
% \PyComment{gallery images, query images}\\ \vspace{-2pt}
% \hspace{34pt} $f_{\phi}$ \hspace{18pt}
% \PyComment{pre-trained embedding network}\\ \vspace{4pt}
% %
% \hspace{-3pt}\vspace{3pt}
% \PyComment{extract features}\\ \vspace{1pt}
% %
% \hspace{-3pt}\vspace{3pt}
% $\mathbf{f_g}=f_{\phi}(\mathbf{x_g}), \:\mathbf{f_q}=f_{\phi}(\mathbf{x_q})$ \\ \vspace{2pt}
% %
% \hspace{-3pt}
% \PyComment{transform features using BPA}\\\vspace{3pt}
% %
% \hspace{-3pt}
% $(\mathbf{f_g}\cup\mathbf{f_q})=\:$\PyCode{{\color{purple}{BPA}}}\hspace{1pt}$(\mathbf{f_g}\cup\mathbf{f_q})$\\\vspace{3pt}
% %
% \hspace{-3pt}
% \PyComment{retrieve gallery image with closest feature}\\\vspace{3pt}
% %
% \hspace{-3pt}
% \textbf{return} $\underset{\{j:\mathbf{f}_j\in \mathbf{f_g}\}}{\mathrm{argmin}}$ $\|\mathbf{f}_i-\mathbf{f}_j\|$ for every $\{i:\mathbf{f}_i\in \mathbf{f_q}\}$ $\hspace{50pt}$\\
% %
% \end{algorithm}
% % }
% % ======================== TDB ================= %%%%
%%%%%%%%%%%%%%%%%%%%%%%%%%%%%%%%%%%%%
%\newpage

% \begin{figure*}[]
% \hspace{-8pt}
%  \centering
% \hspace{-5pt}
%  \includegraphics[width=.95.0\textwidth]{figures/synthetic/multi_plots.png} \\% \vspace{10pt} \\\vspace{-11pt}
% (ii) \small{Clustering accuracy across different noise levels $\sigma$ and dimensions $d$. $\;$\textbf{\textbf{Note}:} For each configuration, BPA is shown by a \textit{dashed} line while the baseline features are shown by a \textit{solid} line.}\\
% \vspace{0pt}\caption{\textbf{Clustering on the sphere}: 2] Results.: In can be seen that BPA (dashed) shows superior results in all aspects (see text for explanations and interpretation).
%    %terms of accuracy for a wide range of noise levels as well as when the clustering sampled from a distribution of higher $\sigma$. Surprisingly, BPA also seems to hinder a fast decrease in the accuracy when $d$ increases.
%    }
%     \label{fig.synthetic_exp}\vspace{-2pt}
% \end{figure*}
% %% ---------------------------------------------------

\section{Clustering on the Sphere - a Case Study} \label{sec:case_study}
\begin{figure*}[t!]
\vspace{-8pt}
\hspace{3pt}
 \setlength{\tabcolsep}{-0.32em} % for the horizontal padding
\begin{tabular}{c c c c c}
     \includegraphics[width=0.2066\textwidth]{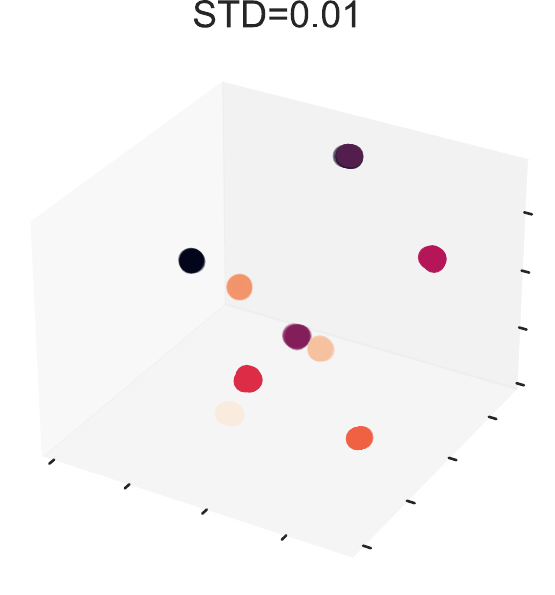} &
     \includegraphics[width=0.2066\textwidth]{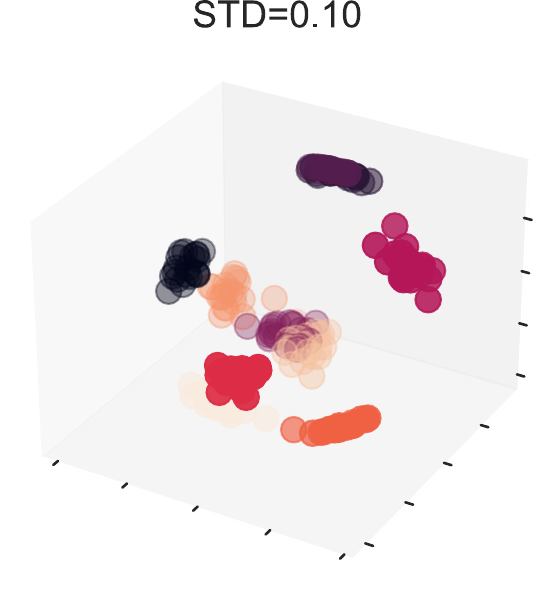}  &
     \includegraphics[width=0.2066\textwidth]{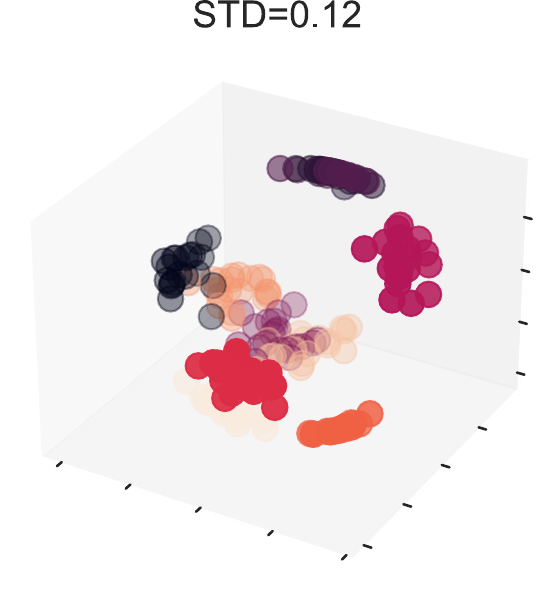} &
     \includegraphics[width=0.2066\textwidth]{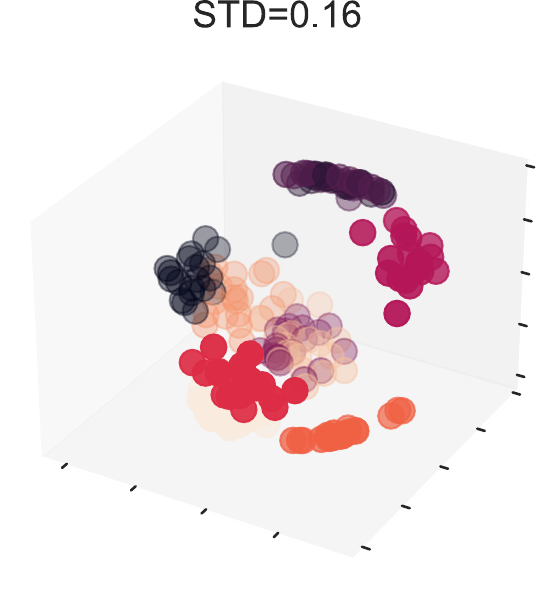} &
     \includegraphics[width=0.2066\textwidth]{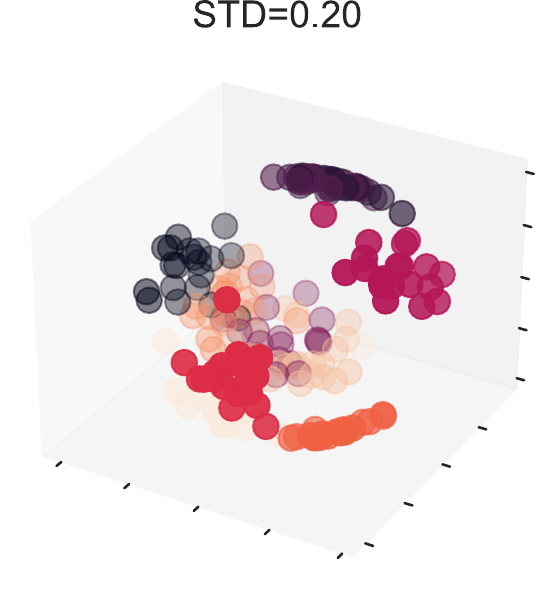} \\
     \includegraphics[width=0.2066\textwidth]{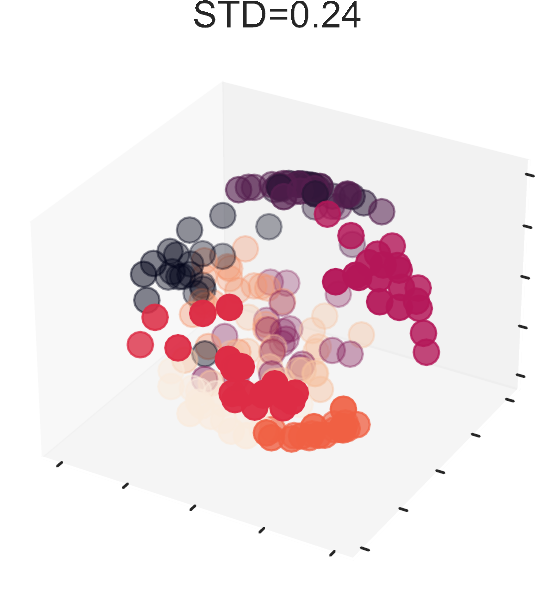} &
     \includegraphics[width=0.2066\textwidth]{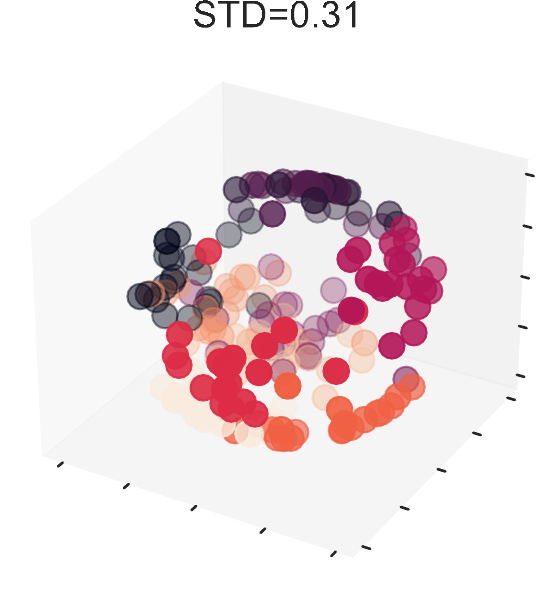} &     \includegraphics[width=0.2066\textwidth]{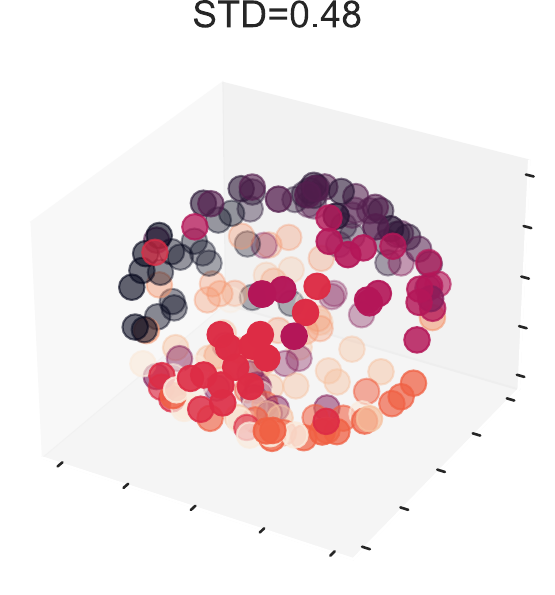} &
     \includegraphics[width=0.2066\textwidth]{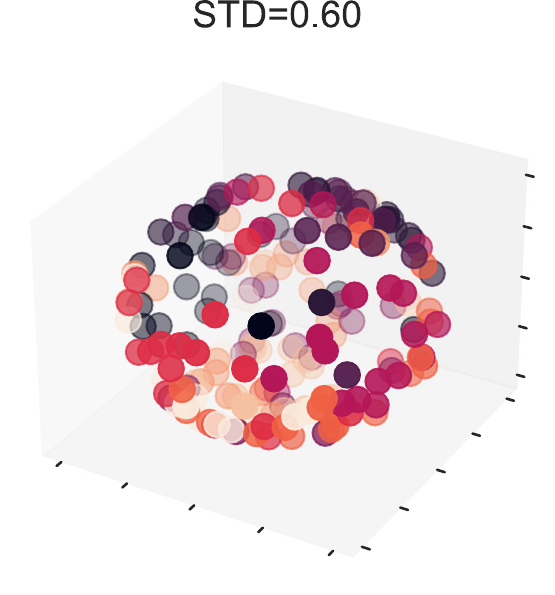}  &
     \includegraphics[width=0.2066\textwidth]{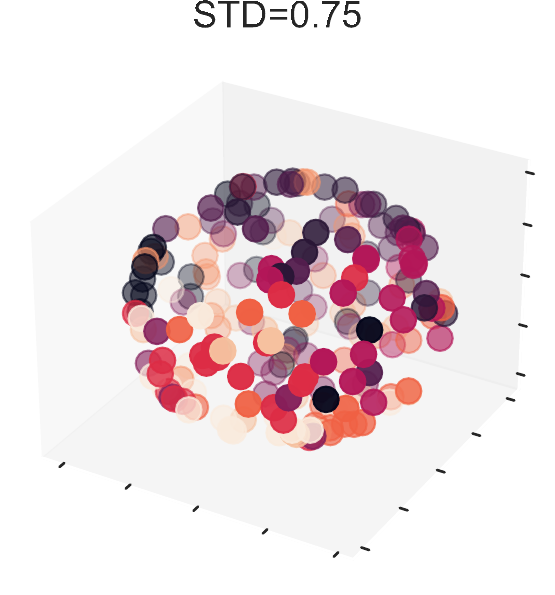} \\\vspace{-12pt}
\end{tabular}\\
\vspace{-13pt}\caption{\textbf{Clustering on the sphere: Data Generation.}  10 Random cluster centers on the unit sphere, perturbed by increasing noise STD.} 
    \label{fig.synthetic_exp:setup}\vspace{-8pt}
\end{figure*}

We demonstrate the effectiveness of BPA using a \underline{controlled synthetically generated} clustering experiment, with $k=10$ cluster centers that are distributed uniformly at random on a $d$-dimensional unit-sphere, and 20 points per cluster (200 in total) that are perturbed around the cluster centers by Gaussian noise of increasing standard deviation, of up to 0.75, followed by a re-projection back to the sphere by dividing each vector by its $L_2$ magnitude.
See Fig. \ref{fig.synthetic_exp:setup} for a visualization of the 3$D$ case, for several noise STDs. Following the random data generation, we also apply dimensionality reduction with PCA to $d=50$, if $d>50$.

%% -------------------- figure SYNTHETIC ---------------
%--------------------------------------------------------

We performed the experiment over a logarithmic 2D grid of combinations of data dimensionalities $d$ in the range $[10, 1234]$ and Gaussian in-cluster noise STD in the range $[0.1,0.75]$. 
% Refer to Fig.~\ref{fig.synthetic_exp} (i) for a visualization of the data generation process. 
Each point is represented by its $d$-dimensional coordinates vector, where the baseline clustering is obtained by running k-means on these location features. In addition, we run k-means on the set of features that has undergone BPA. Hence, the benefits of the transform (embedding) are measured indirectly through the accuracy\footnote{Accuracy is measured by comparison with the optimal permutation of the predicted labels, found by the Hungarian Algorithm \cite{kuhn1955hungarian}.} achieved by running k-means on the embedded vs. original vectors.

%% -------------------- figure SYNTHETIC ---------------
\begin{figure*}[b]
\centering\vspace{-6pt}
 \includegraphics[width=0.90\textwidth]{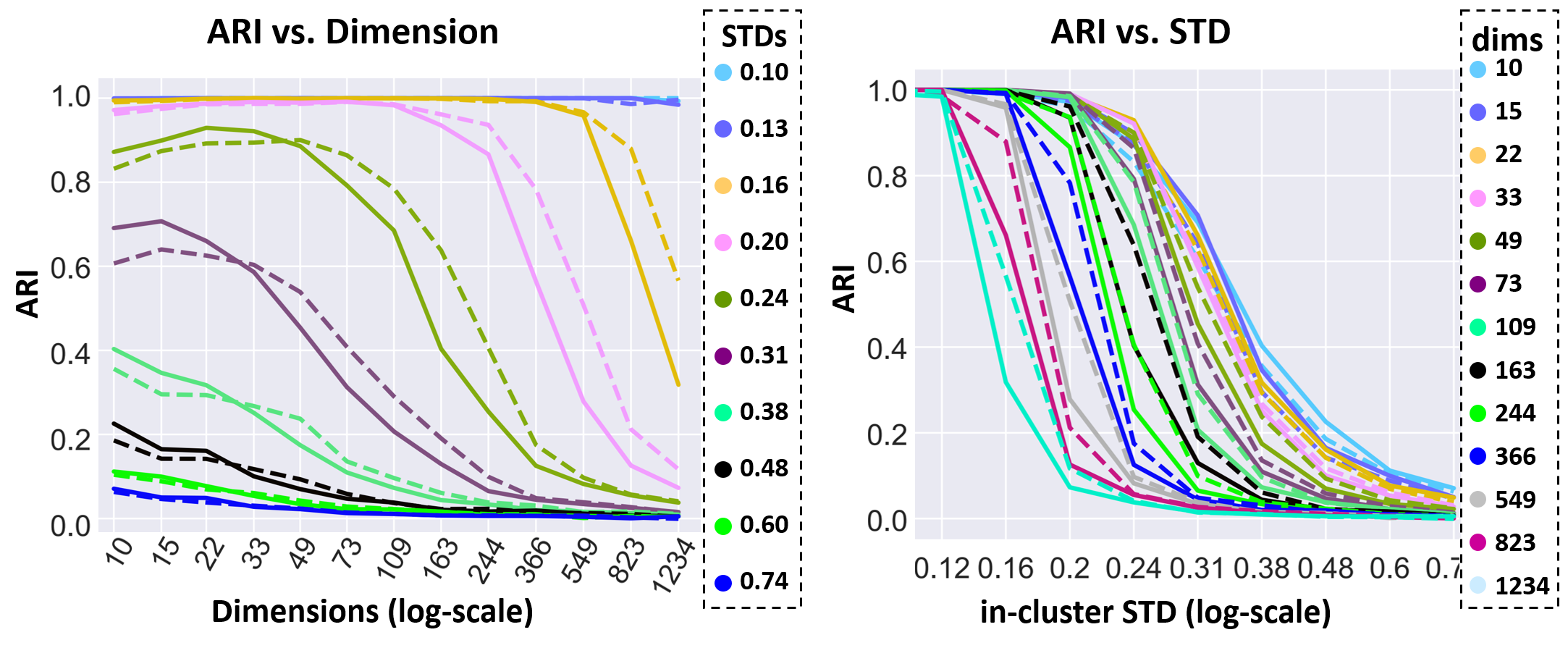} \\ \vspace{2pt}
 \includegraphics[width=0.90\textwidth]{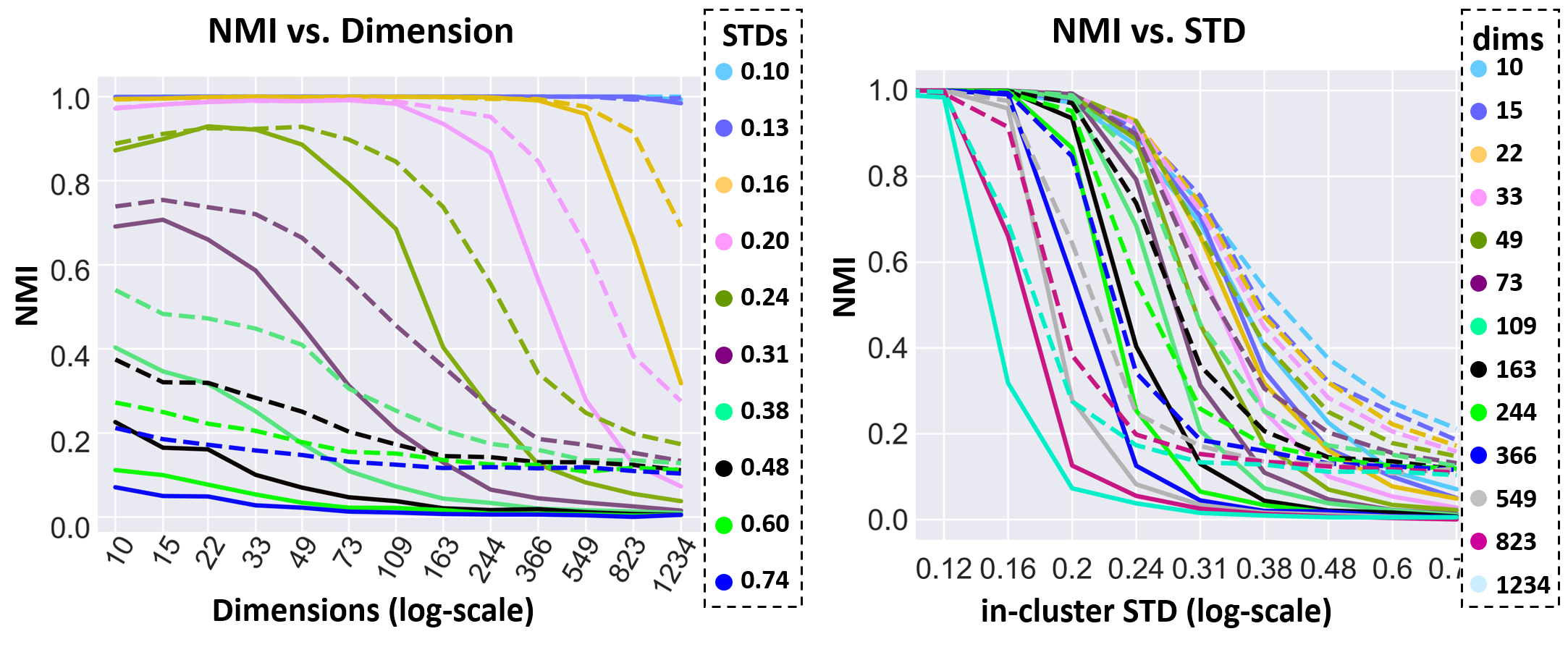} \\\vspace{-8pt}\caption{\textbf{Clustering on the sphere: Detailed Results.} 
  Clustering measures (top: ARI, bottom: NMI) of $k$-means, using BPA features (\textit{dashed} lines) vs. original features (\textit{solid} lines). For both measures - the higher the better. Shown over different configurations of feature dimensions $d$ (left) and noise levels $\sigma$ (right).}
    \label{fig.synthetic_exp}\vspace{-22pt}
\end{figure*}
%% ---------------------------------------------------

Evaluation results, in terms of Normalized Mutual Information (NMI) and Adjusted Rand Index (ARI), are reported in Fig.~\ref{fig.synthetic_exp}, averaged over 10 runs, as a function of either dimensionality (for different noise STDs) or noise STDs (for different dimensionalities). The results show (i) general  gains and robustness to wide ranges of data dimensionality (ii) the ability of BPA to find meaningful representations that enable clustering quality to degrade gracefully with the increase in cluster noise level. Note that the levels of noise are rather high, as they are relative to a unit radius sphere. 

%%%%%%%%%%%%%%%%%%%%%%%%%%%%%%%%%%%%%%%%%%%%%%%%%%%%%%%%%%%%%%%%%%%%%%%%%%%%%%%
%%%%%%%%%%%%%%%%%%%%%%%%%%%%%%%%%%%%%%%%%%%%%%%%%%%%%%%%%%%%%%%%%%%%%%%%%%%%%%%

\end{document}